%% file: main.tex
\documentclass[conference]{IEEEtran}
\usepackage{times}
\usepackage[numbers]{natbib}
\usepackage{multicol}
\usepackage{booktabs}
\usepackage[bookmarks=true]{hyperref}
\usepackage{amsmath}
\usepackage{amssymb}
\usepackage{graphicx}
\usepackage{subcaption}

\input{preamble}

\input{math}
\begin{document}
\definecolor{myred}{RGB}{254,113,52}

\title{\textcolor{myred}{HAIC}: \textcolor{myred}{H}umanoid \textcolor{myred}{A}gile Object \textcolor{myred}{I}nteraction \textcolor{myred}{C}ontrol via Dynamics-Aware World Model}
\author{
Dongting Li\textsuperscript{1,4}$^*$, 
Xingyu Chen\textsuperscript{2,4}$^*$, 
Qianyang Wu\textsuperscript{4}, 
Bo Chen\textsuperscript{4}, 
Sikai Wu\textsuperscript{4},
Hanyu Wu\textsuperscript{3},
Guoyao Zhang\textsuperscript{4}\\

Liang Li\textsuperscript{4},
Mingliang Zhou\textsuperscript{4},
Diyun Xiang\textsuperscript{4},
Jianzhu Ma\textsuperscript{1}\authorrefmark{2}, 
Qiang Zhang\textsuperscript{2}\authorrefmark{2}, 
Renjing Xu\textsuperscript{2}\authorrefmark{2}\\

\authorblockA{
\textsuperscript{1}Tsinghua University\quad 
\textsuperscript{2}HKUST(Guangzhou)\quad 
\textsuperscript{3}ETH Zurich\quad 
\textsuperscript{4}Xiaomi Robotics Lab\\
\small{$^*$Equal Contribution \quad \authorrefmark{2}Equal Advising}\\
\href{https://haic-humanoid.github.io/}{\texttt{https://haic-humanoid.github.io/}}
}
}

\twocolumn[{%
\renewcommand\twocolumn[1][]{#1}%
\maketitle
\vspace{-0.45cm}

\begin{center}
    \centering
    \captionsetup{type=figure}
    \includegraphics[width=1.0\textwidth]{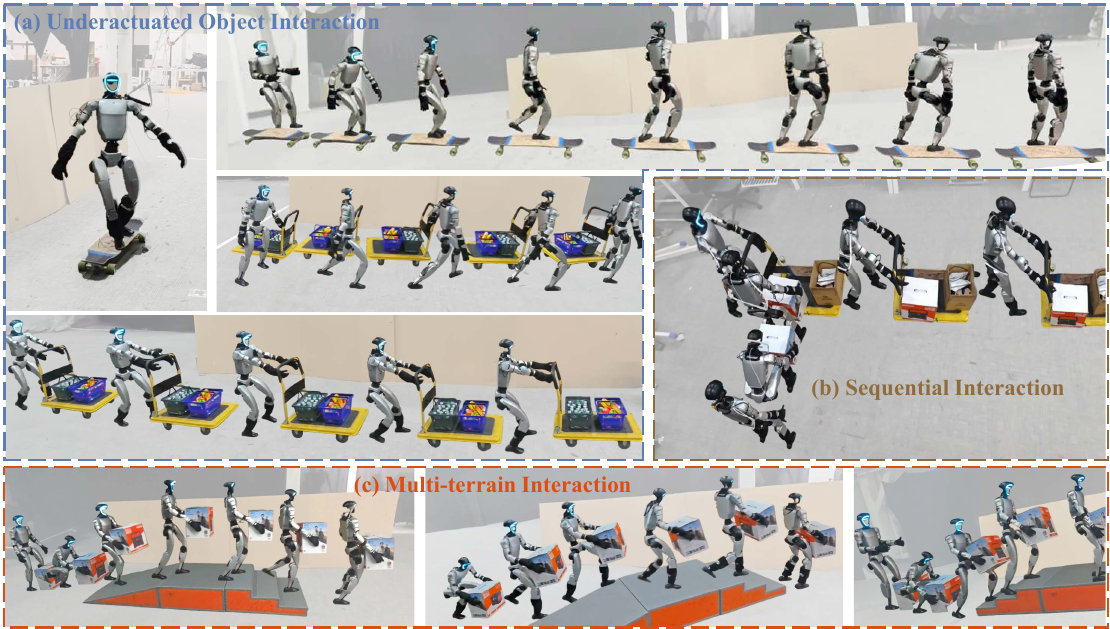}
    \vspace{-0.17in}
    \caption{Our proposed framework \ours enables a robot to perform complex interactions, including (a) \textbf{Underactuated Object Interaction.} The robot learns interaction skills such as skateboarding, cart pushing, and cart pulling. (b) \textbf{Sequential Interaction.} \ours supports multi-object interaction, enabling a whole-body controller to pick up the box, load it onto the cart, then drive them forward in one policy. (c) \textbf{Multi-terrain Interaction.} \ours enables robots to maintain interactions with both an object and the scene, such as carrying a box across various types of terrains, such as platforms, slopes, and stairs.} 
    \label{fig:teaser}
\end{center}
\vspace{0.04in}
}]

\begin{abstract}

Humanoid robots exhibit significant potential for executing complex whole-body interaction tasks in unstructured environments. While recent advancements in Human-Object Interaction (HOI) have been substantial, prevailing methodologies predominantly address the manipulation of fully actuated objects, where the target is rigidly coupled to the robot's end-effector and its state is strictly constrained by the robot's kinematics. This paradigm neglects the pervasive class of underactuated objects characterized by independent dynamics and non-holonomic constraints, which pose significant control challenges due to complex coupling forces and frequent visual occlusions. 
To bridge this gap, we propose \ours, a unified framework designed to enable robust interaction across a spectrum of object dynamics without reliance on external state estimation. Central to our approach is a novel dynamics predictor that infers high-order object states, specifically velocity and acceleration, solely from proprioceptive history. These predictions are explicitly projected onto static geometric priors to construct a spatially grounded representation of dynamic occupancy, allowing the policy to internalize collision boundaries and contact affordances in visual blind spots. We employ an asymmetric fine-tuning strategy where the world model continuously adapts to the student policy’s exploration, ensuring robust state estimation under distribution shifts. 
We evaluate our framework on a Unitree G1 humanoid robot. Empirical results demonstrate that \ours achieves high success rates in agile object interactions, including skateboarding, cart pushing, and cart pulling under various weight load conditions, by compensating for inertial physical perturbations. Furthermore, by accurately predicting multi-object dynamics, it seamlessly masters sequential tasks and carries a box across composed terrains.

\end{abstract}

\IEEEpeerreviewmaketitle

\input{sections/1_introduction}
\input{sections/2_related_work}

\input{sections/3_method}
\input{sections/4_experiments}
\input{sections/5_conclusion}


\bibliographystyle{plainnat}
\bibliography{references}

\clearpage
\input{sections/appendix}

\end{document}

%% file: preamble.tex
\usepackage{multirow}
\usepackage{multicol}
\usepackage{graphicx}
\usepackage{xspace}
\usepackage{xcolor}
\usepackage{caption}
\usepackage{wrapfig}
\usepackage{bbding}  
\usepackage{pifont}  

\usepackage{booktabs}
\usepackage{float}
\usepackage{amsmath}  
\usepackage{amssymb}  
\usepackage{bm}
\usepackage{bbm}
\usepackage{dsfont}
\usepackage{makecell}
\newcommand{\ours}[0]{\textsc{HAIC}\xspace}
\newcommand{\ma}[0]{{Proprio}\xspace}
\newcommand{\mb}[0]{{Vec-Pose}\xspace}
\newcommand{\mc}[0]{{Vec-Dyn}\xspace}
\newcommand{\md}[0]{{Geo-Pose}\xspace}
\newcommand{\baseline}[0]{{HDMI$^*$}\xspace}

\usepackage{hyperref}
\usepackage[capitalise, nameinlink]{cleveref}

\usepackage{colortbl}
\definecolor{ourcolor}{HTML}{99e0eb}
\definecolor{ourblue}{HTML}{27a2c3}

\definecolor{tablecolor}{HTML}{ccf2f5} 

\definecolor{tablecolor2}{HTML}{ffcdb4}
\definecolor{citecolor}{HTML}{fe7b5b}
\definecolor{grey}{rgb}{0.9, 0.9, 0.9}
\usepackage{amssymb}

\usepackage{listings}
\definecolor{gred}{rgb}{0.859,0.267,0.216}
\definecolor{ggreen}{rgb}{0.059,0.616,0.345}

\definecolor{deepblue}{HTML}{27a2c3}

\definecolor{deepred}{HTML}{fe7b5b}

\usepackage[font=small,labelfont=bf]{caption}

\usepackage[font=footnotesize,labelfont=bf]{caption}

\definecolor{citecolor}{HTML}{faa700} 
\definecolor{lblue}{HTML}{ffb114} 
\definecolor{ogreen}{HTML}{2E7D32}
\definecolor{bred}{HTML}{BF360C}
\definecolor{newbrown}{HTML}{795548}

\hypersetup{
    colorlinks=true,
    linkcolor=brown,
    filecolor=magenta,      
    urlcolor=brown,
    citecolor=brown,
}

\usepackage{multicol}
\usepackage{multirow}
\usepackage{colortbl}
\usepackage{booktabs}   
\usepackage{bbding} 
\usepackage{graphicx}
\let\labelindent\relax 
\usepackage{enumitem}

\usepackage{bbding}

\usepackage{pifont}
\usepackage{xfrac}
\usepackage{arydshln}

%% file: sections/1_introduction.tex
\section{Introduction}

Humanoid robots, characterized by their anthropomorphic morphology, possess immense potential for executing complex whole-body interaction tasks in unstructured environments. From transporting payloads to interacting dynamically with tools, these tasks require not only robust locomotion stability but also the capacity to perform precise interactions with the environment. Recent data-driven approaches in Human-Object Interaction have achieved significant milestones, enabling robots to acquire diverse interaction skills~\citep{he2024hover,dao2024sim,liu2024opt2skill,wang2024hypermotion}.

However, prevailing research predominantly focuses on the Interaction of \textbf{fully actuated objects}, such as carrying a rigidly grasped box~\citep{dao2024sim,liu2024opt2skill}. In these scenarios, the object's state is strictly constrained by the robot's end-effector, resulting in relatively simple and fully controlled dynamics. In contrast, the real world is populated with \textbf{underactuated objects} possessing independent dynamics and non-holonomic constraints, such as skateboards, carts, or luggage. Manipulating such objects presents a dual challenge. First, \textbf{visual occlusion} is frequent. Large objects often obstruct onboard sensors, depriving the robot of direct observation of both the terrain and the object itself~\citep{Lauri_2023}. Second, \textbf{dynamics coupling} becomes critical. Unlike static payloads, underactuated objects generate independent acceleration and inertial forces that propagate to the robot. Traditional control methods often rely on external high-precision state estimation to handle these disturbances, which is impractical for real-world deployment~\citep{sipos2022simultaneouscontactlocationobject}. Meanwhile, existing reinforcement learning policies, lacking explicit modeling of high-order dynamics, often suffer from response lag, leading to instability during high-speed interactions~\citep{hwangbo2019learning,li2024reinforcement,radosavovic2024real}. 

Humans exhibit remarkable adaptability when facing similar constraints. Even when vision is obstructed by a carried object or when balancing on an accelerating skateboard, humans rely on proprioception and an internal mental model to infer the object's state. This capability stems from a fusion of geometric reasoning (knowing the object's boundaries) and dynamic sensing (feeling the forces to anticipate acceleration), enabling proactive center-of-mass adjustments.
\begin{figure}[t]
    \centering
    \begin{subfigure}[b]{0.38\linewidth}
        \centering
        \includegraphics[width=\linewidth]{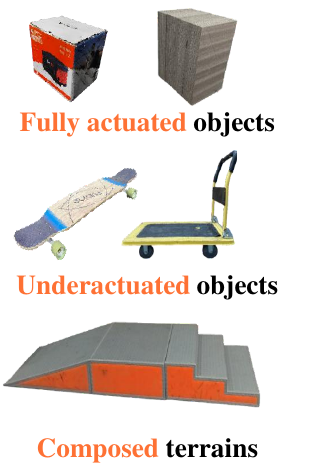}
        \caption{Types of interacted objects.} 
        \label{fig:objects_intro}
    \end{subfigure}
    \hfill
    \begin{subfigure}[b]{0.60\linewidth}
        \centering
        \includegraphics[width=\linewidth]{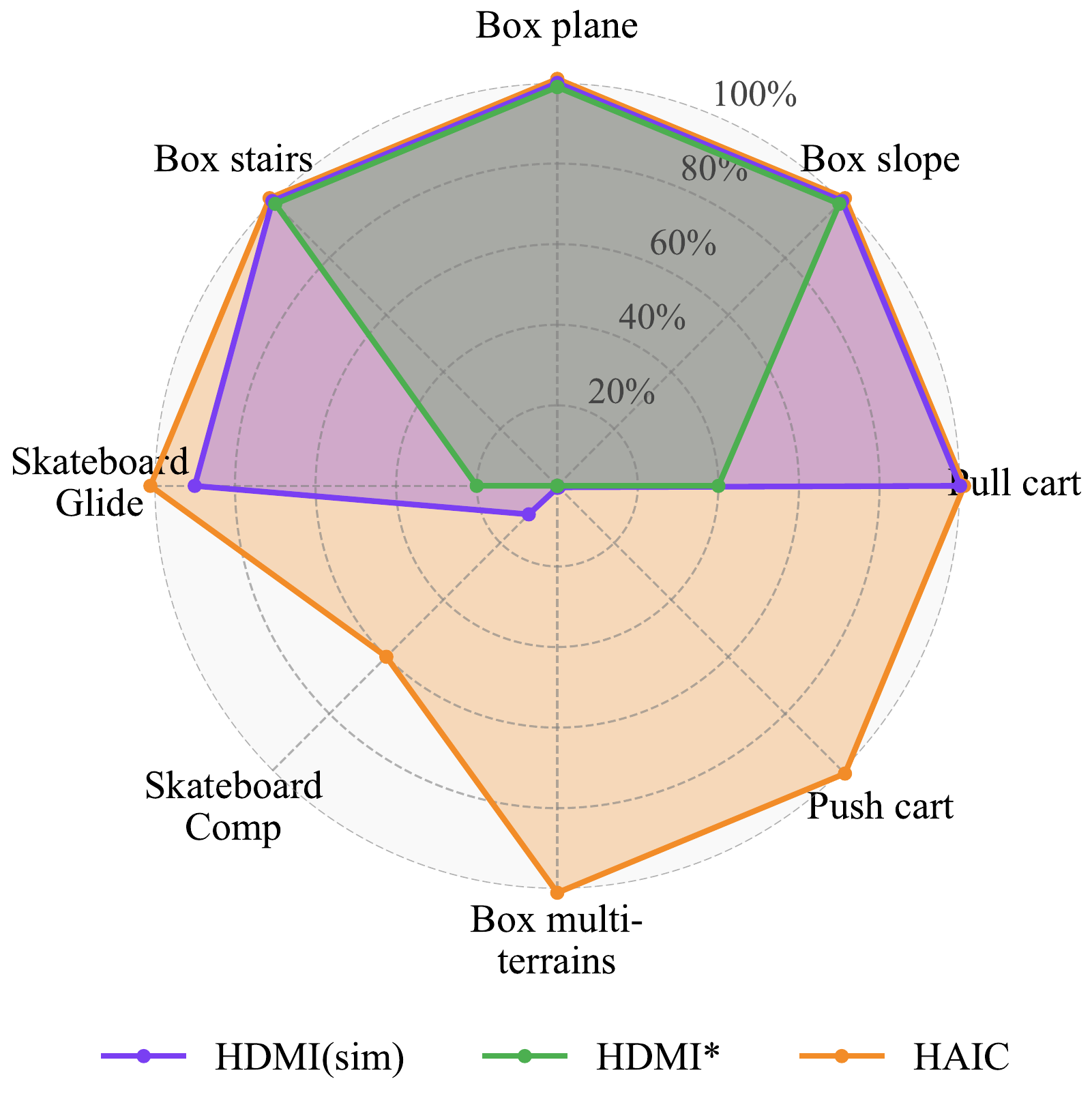}
        \caption{Success rate comparison on different tasks.} 
        \label{fig:radar_comparison}
    \end{subfigure}
    \caption{\ours excels at complex interactions, particularly with underactuated objects, and significantly outperforms the baseline.} 
    \label{fig:intro_main}
    \vspace{-0.2in}
\end{figure}

Inspired by this cognitive mechanism, we introduce \ours, a unified learning framework designed to enable robust and agile object interaction in visual blind spots. Unlike prior models focused on visual prediction~\citep{ha2018world,hafner2023mastering,hansen2024hierarchical,gu2024advancing,long2024learning}, \ours is proprioception-centric and addresses the control of both fully actuated and underactuated objects through two key mechanisms in a dynamics-aware world model (DWM):

\begin{itemize}

    \item High-Order Dynamics Prediction: By analyzing proprioceptive history, \ours explicitly predicts the object's linear and angular accelerations. This allows the robot to capture the independent motion trends of underactuated objects, facilitating proactive inertia compensation.
    \item Explicit Geometric Projection: \ours explicitly project these predicted dynamic states onto a static geometric prior (point cloud) of the object. This constructs a spatially grounded representation of dynamic occupancy, enabling the policy to perceive collision boundaries and contact affordances even without visual feedback.

\end{itemize}

To bridge the gap between simulation and reality, we employ an asymmetric fine-tuning strategy where the world model continuously adapts to the student policy's exploration online, effectively mitigating distribution shifts~\citep{pinto2017asymmetricactorcriticimagebased,tobin2017domain}.

The main contributions of this work are summarized as follows:
\begin{itemize}
    \item \textbf{Dynamics-Aware World Model:} We propose a novel dynamics predictor that integrates high-order acceleration inference with explicit geometric projection, unifying state estimation for both static payloads and dynamic underactuated objects.
    \item \textbf{Asymmetric Adaptive Distillation:} We design a robust two-stage training pipeline that transfers privileged geometric and dynamic reasoning into a sensor-limited student policy via continuous world model adaptation.
    \item \textbf{Agile Real-World Performance:} We demonstrate SOTA performance on a Unitree G1 humanoid robot. Our framework achieves a 100\% success rate in blind multi-terrain box carrying and, for the first time without external sensing, enables the robot to master agile underactuated tasks, such as dynamic skateboarding.
\end{itemize}

%% file: sections/2_related_work.tex
\section{Related Work}

\subsection{Humanoid Whole-Body Control and Tracking}
Deep reinforcement learning (RL)~\cite{schulman2017proximalpolicyoptimizationalgorithms} has achieved significant milestones in tracking large-scale human motion capture (MoCap) datasets like AMASS~\cite{mahmood2019amassarchivemotioncapture}. Recent General Motion Trackers (GMT)~\cite{chen2025gmt, yin2025unitrackerlearninguniversalwholebody} and retargeting frameworks~\cite{joao2025gmr, liao2025beyondmimicmotiontrackingversatile, he2024hover} allow humanoids to reproduce diverse kinematic behaviors. To support these data-driven approaches, scalable collection systems like TWIST~\cite{ze2025twist} and TWIST2~\cite{ze2025twist2} have been developed to provide high-quality whole-body demonstrations.


Early studies demonstrated that motion imitation and adversarial motion priors enable policies to track diverse human motion data, producing natural and robust whole-body control \cite{peng2018deepmimic, peng2021amp, xie2025kungfubot,  liao2025beyondmimicmotiontrackingversatile, he2025asap, huang2025learning}. These works establish motion tracking as a scalable paradigm for acquiring rich motor skills while substantially reducing manual reward engineering.

Building upon this foundation, subsequent research \cite{ze2025twist2, han2025kungfubot2,yuan2025behavior, zeng2025behavior} investigates generalized whole-body control through large-scale motion datasets and unified latent representations, enabling a single policy to compose heterogeneous skills and adapt to diverse task conditions. Policies are pretrained via large-scale data and unsupervised reinforcement learning, and conditioned on language, vision, or high-level task descriptors to enable controllable and zero-shot skill adaptation \cite{li2025bfm, luo2025sonic, jiang2025uniact, shao2025langwbc}. These models demonstrate strong generalization across diverse motion patterns, highlighting the potential of foundation training for whole-body control and advanced interactions with environments, especially dynamic objects.


\subsection{Physics-Based Human-Object Interaction}
Humanoid-object interaction (HOI) is a longstanding challenge bridging computer vision and robotics. Large-scale HOI datasets provide diverse demonstrations to facilitate data-driven learning~\cite{li2023object,taheri2020grab,bhatnagar2022behave,jiang2023full,zhang2023neuraldome,xu2025interact}. Based on these data, advanced frameworks~\cite{wang2023physhoi,tessler2024maskedmimic,Wang_2025_CVPR,pan2025tokenhsi,xu2025intermimic,lin2025simgenhoiphysicallyrealisticwholebody} have been developed to train policies that master versatile whole-body interactions in simulation. Similarly, video-based imitation methods extract interactive skills directly from human videos \cite{weng2025hdmilearninginteractivehumanoid,allshire2025videomimic,shao2025visualmimic}. Prior studies on loco-manipulation, which focus only on the interaction between the hand and objects, show that integrating force adaptation \cite{zhang2025falcon}, hierarchical control \cite{li2025learning, kalaria2025dreamcontrol, li2025amo}, or residual action \cite{sun2025ulc} enables precise manipulation behaviors with stable locomotion. And visual sim-to-real transfer methods substantially improve real-world generalization for continuous manipulation tasks \cite{xue2025leverb, he2025viral, jiang2025wholebodyvlaunifiedlatentvla}. Some specialized systems demonstrate high-speed adaptation for specific sports but lack generality~\cite{su2025hitterhumanoidtabletennis, xu2025learning, liu2025humanoid,wang2025learning}. As for full-body interactions, Omniretarget~\cite{yang2025omniretarget} improves the quality of retargeted motion by constructing the interaction mesh. HDMI~\cite{weng2025hdmilearninginteractivehumanoid} and ResMimic~\cite{zhao2025resmimicgeneralmotiontracking} incorporate residual action modules to adapt policies to virtual object states. Other methods typically rely on explicit object tracking to estimate object states, such as PhysHSI (LiDAR)~\cite{wang2025physhsi}, HDMI (markers)~\cite{weng2025hdmilearninginteractivehumanoid}, VisualMimic (depth)~\cite{shao2025visualmimic}, and DemoHLM (RGB)~\cite{fu2025demohlm}. 
However, all these methods are restricted to interactions with static environments or fully actuated objects. In contrast, \ours can learn coupled humanoid-object dynamics through a specialized world model, enabling agile interactions with underactuated objects or multi-terrain environments.

\subsection{World Models for Humanoids}

World models \cite{ha2018world, hafner2023mastering} act as internal simulators that enable agents to anticipate future outcomes, and have been widely adopted in robotic manipulation to predict tactile feedback and object dynamics\cite{feng2025learninginteractiveworldmodel,jeong2025objectcentricworldmodellanguageguided, sipos2022simultaneouscontactlocationobject}.
Recent work leverages world models to address high-dimensional humanoid dynamics and complex contact sequences. DWL~\cite{gu2024advancing} reconstructs complete physical states from noisy observations via denoising learning, enabling robust whole-body control, while RWM~\cite{li2025robotic} mitigates long-horizon drift through a dual-autoregressive prediction scheme. WMR~\cite{sun2025learning} decouples dynamics learning from policy gradients and enforces state reconstruction to prevent representation collapse. World models are also used as self-supervised objectives for perceptual robustness, as demonstrated by WMP~\cite{lai2025world}, and extended to active interaction in Ego-VCP~\cite{liu2025ego} for ego-centric contact planning. Any2Track~\cite{zhang2025track} learns dynamics-aware embeddings from observation histories, enabling implicit inference of environmental properties and robust sim-to-real transfer. However, these methods primarily model proprioceptive dynamics for whole-body control, often neglecting the coupled dynamics inherent in object interaction.

%% file: sections/3_method.tex
\section{Methodology}
\begin{figure}[t]
    \centering
    \includegraphics[width=\linewidth]{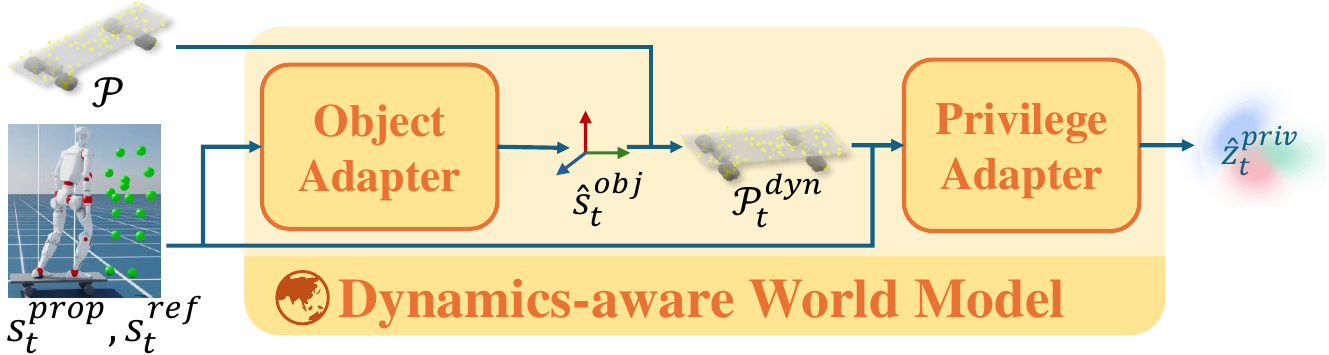}
    \caption{\textbf{Overview of our Dynamics-aware World Model.} It predicts object dynamics from proprioception and explicitly projects them onto a static geometric prior for reconstructing the privileged information.} 
    \label{fig:world_model}
    \vspace{-0.2in}
\end{figure}
\subsection{Problem Formulation}
The purpose of humanoid-object interaction is to train a whole-body control policy that simultaneously tracks the reference trajectory of both the robot and the object. We formulate this proprioception-based interaction task as a partially observable Markov decision process (POMDP)~\cite{Lauri_2023}. At each timestep $t$, the full state $s_t \in \mathcal{S}$ includes the robot's proprioception $s^{prop}_t$, the reference motion $s^{ref}_t$, the point cloud of objects $\mathcal{P}$ and the privileged state $s^{priv}_t$.

Since the object state $s^{obj}_t$ cannot be acquired directly, we adapt the teacher-student framework~\cite{kumar2021rma} to distill the privileged information. In training, the teacher has access to the full state $s_t$. During deployment, the student only observes:
\begin{itemize}
    \item Onboard proprioception $s^{prop}_t$ (joint data, IMU data).
    \item Reference motion $s^{ref}_t$.
    \item Nominal geometric template $\mathcal{P} \in \mathbb{R}^{N \times 3}$: a standard-sized point cloud acting as a semantic prior.
\end{itemize}
The goal is to explore a policy $\pi(a_t | s^{prop}_t, s^{ref}_t, a_{t-1}, \mathcal{P})$ in the simulation that enables the humanoid robot to acquire robust interaction skills with objects in the real world. 

\begin{figure*}[t]
    \centering
    \includegraphics[width=\linewidth]{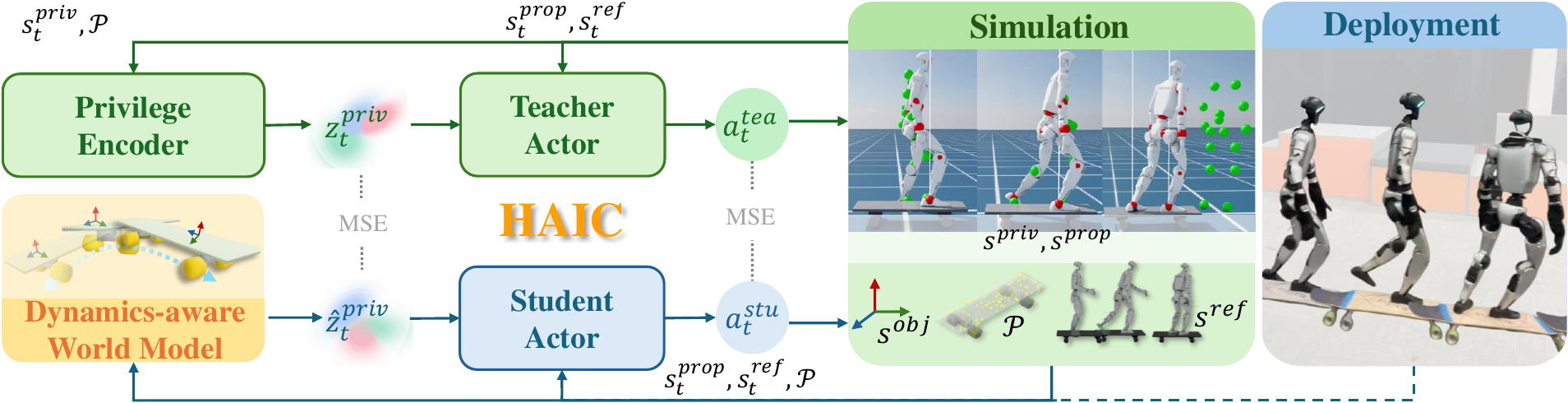}
    \caption{\textbf{Framework overview}. We train policies in the simulation from scratch. The framework includes a privileged teacher and a dynamics-aware student. The student policy utilizes the learned world model to perform robust interaction tasks such as skateboarding on a real humanoid robot.} 
    \label{fig:framework}
    \vspace{-0.2in}
\end{figure*}

\subsection{Dynamics-Aware World Model}
Our motivation is to use the robot's state $s^{prop}_t$ to predict the inaccessible state of objects $s^{obj}_t$~\cite{sipos2022simultaneouscontactlocationobject}. Therefore, we design a predictive framework to internalize the object's dynamics relative to the robot. And we propose a pipeline that first estimates object dynamics, explicitly projects these dynamics onto the geometric prior, and subsequently feeds them into the downstream network, as shown in Fig.~\ref{fig:world_model}.

\subsubsection{Object Adapter}
The core component of the dynamics-aware world model is the object adapter, which functions as a real-time dynamics predictor. It takes the history of predicted object state $\hat{s}^{obj}_{t-1}$, robot observations $\mathcal{H}_t = \{s^{prop}_{t-n}, \dots, s^{prop}_t\}$ and the future of reference motion $\mathcal{F}_t = \{s^{ref}_{t}, \dots, s^{ref}_{t+m}\}$ as input and predicts the object's relative spatial and temporal state $\hat{s}^{obj}_{t}$:
\begin{equation}
    \hat{s}^{obj}_t = f_{OA}(\hat{s}^{obj}_{t-1}, \mathcal{H}_t, \mathcal{F}_t)
\end{equation}
where $\hat{s}^{obj}_t$ includes the predicted relative position $\hat{p}_t$, orientation $\hat{R}_t$, linear/angular velocities ($\hat{v}^{lin}_t, \hat{v}^{ang}_t$), and linear/angular accelerations ($\hat{a}^{lin}_t, \hat{a}^{ang}_t$). 

\subsubsection{Explicit Geometric Projection}
While dynamics-aware approaches using signed distance fields (SDFs) or point clouds have been explored for stationary arms~\cite{Li2024OKAMITH, jeong2025objectcentricworldmodellanguageguided} and HOI animation~\cite{al2013relationship, wu2024thor}, the application to the high-dimensional dynamics of humanoid whole-body control remains limited. By conditioning our predictor on a point-cloud-based geometric prior, the robot can mentally track an invisible object, effectively transforming a partially observable Markov decision process into a manageable control task without the need for the real-time privileged states.

To bridge the gap between abstract state estimation and physical interaction, we introduce an explicit geometric projection step. We transform the canonical point cloud $\mathcal{P}$ using the predicted $\hat{s}^{obj}_t$ into the robot's root coordinate frame:
\begin{equation}
    \mathcal{P}^{dyn}_t = \hat{R}_t \cdot \mathcal{P} + \hat{p}_t
    \label{eq:geo_proj}
\end{equation}
By explicitly generating the {dynamic object point cloud} $\mathcal{P}^{dyn}_t$, we provide the privilege adapter with a spatially grounded representation of the object's occupancy. Note that $\mathcal{P}$ serves as a geometric prior rather than a ground-truth measurement.

\subsubsection{Privilege Adapter}
Finally, the estimated dynamic geometry is fused with proprioception via the privilege adapter. This module takes the dynamic point cloud $\mathcal{P}^{dyn}_t$, current proprioception $s^{prop}_t$, and the reference motion $s^{ref}_t$ to produce a predicted privileged feature $\hat{z}^{priv}_t$. This feature serves as the input to the student actor, effectively mimicking the information flow of a privileged actor.

\subsection{Two-Stage Training Framework}
Fig.~\ref{fig:framework} illustrates the overall architecture of \ours. To streamline our formulation, we define the unified {world model loss} ($\mathcal{L}_{WM}$) as the weighted sum of dynamics prediction error and latent feature alignment:
\begin{equation}
    \mathcal{L}_{WM} = \lambda_{\text{obj}} \| \hat{s}^{obj}_t - s^{obj}_t \|^2 + \lambda_{\text{priv}} \| \hat{z}^{priv}_t - z^{priv}_t \|^2
\end{equation}
where $\lambda_{\text{obj}}$ and $\lambda_{\text{priv}}$ balance the supervision strength of physical state regression and latent space distillation, respectively. We denote the standard PPO~\cite{schulman2017proximalpolicyoptimizationalgorithms} objective (including policy gradient, value loss, and entropy) as $\mathcal{L}_{PPO}(\pi, V)$.

\subsubsection{Stage 1: Feature Alignment \& Warm-up}
In the first stage, we train the privileged Teacher actor $\pi_T$ via reinforcement learning. Simultaneously, we train the world model and the student actor $\pi_S$ to mirror the teacher's behavior and internal representation~\cite{pmlr-v15-ross11a}. This ``warm-up'' phase ensures that the student starts Stage 2 with a learned sensorimotor primitive closer to the expert distribution. The objective is:
\begin{equation}
    \mathcal{L}_{Stage1} = \mathcal{L}_{PPO}(\pi_T, V_{T}) + \mathcal{L}_{WM} + \lambda_{\text{distill}} D_{KL}(\pi_S||\pi_T)
\end{equation}
Here, the action distillation term $\lambda_{\text{distill}}$ forces the student to produce actions consistent with the teacher.

\subsubsection{Stage 2: Asymmetric Fine-tuning}
In the second stage, we fine-tune the student actor $\pi_S$ using an asymmetric actor-critic setup~\cite{pinto2017asymmetricactorcriticimagebased}. The critic $V_{T}$ retains access to privileged states for variance reduction. 

To address the distribution shift caused by the student's independent exploration, we continue to optimize the world model alongside the policy. The total objective is:
\begin{equation}
    \mathcal{L}_{Stage2} = \mathcal{L}_{PPO}(\pi_S, V_{T}) + \mathcal{L}_{WM}
\end{equation}
By consistently minimizing $\mathcal{L}_{WM}$ in both stages, we ensure the world model continuously adapts to the evolving trajectory distribution of the student policy, preventing the ``blind'' policy from collapsing due to inaccurate state estimation.

\noindent \textbf{Stabilization via EMA:} To further mitigate instability arising from the coupled training of the policy and the world model, we maintain an exponential moving average (EMA)~\cite{MoralesBrotons2024ExponentialMA} of the world model weights for the rollout policy. The policy interacts with the environment using the EMA-smoothed predictions, while gradients are backpropagated through the online network.

\begin{figure}[ht]
    \centering
    \includegraphics[width=0.8\linewidth]{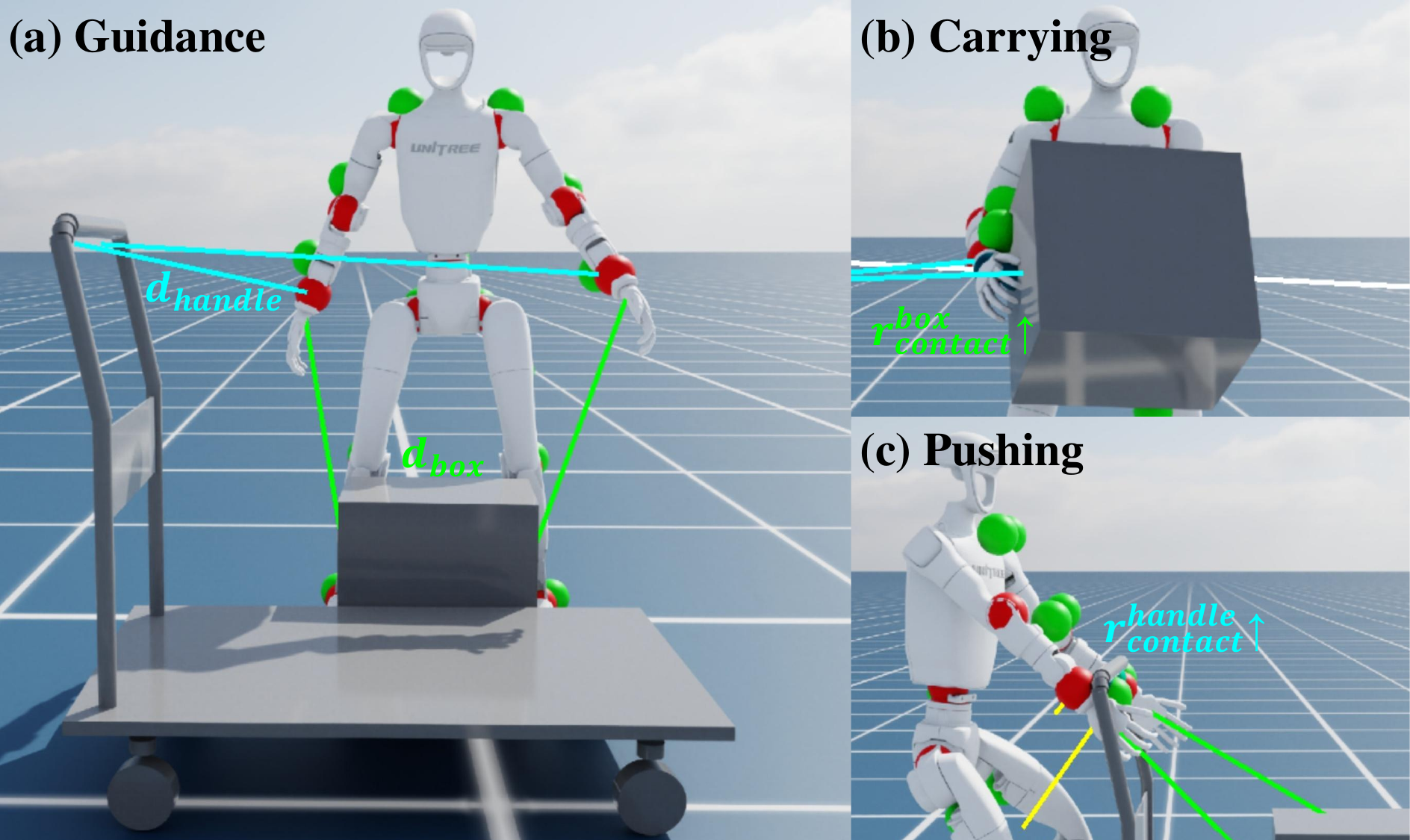}
    \caption{Multiple Objects Contact Guidance Strategy.} 
    \label{fig:contact_reward}
    \vspace{-0.2in}
\end{figure}

\begin{figure*}[ht]
    \centering
    \includegraphics[width=\linewidth]{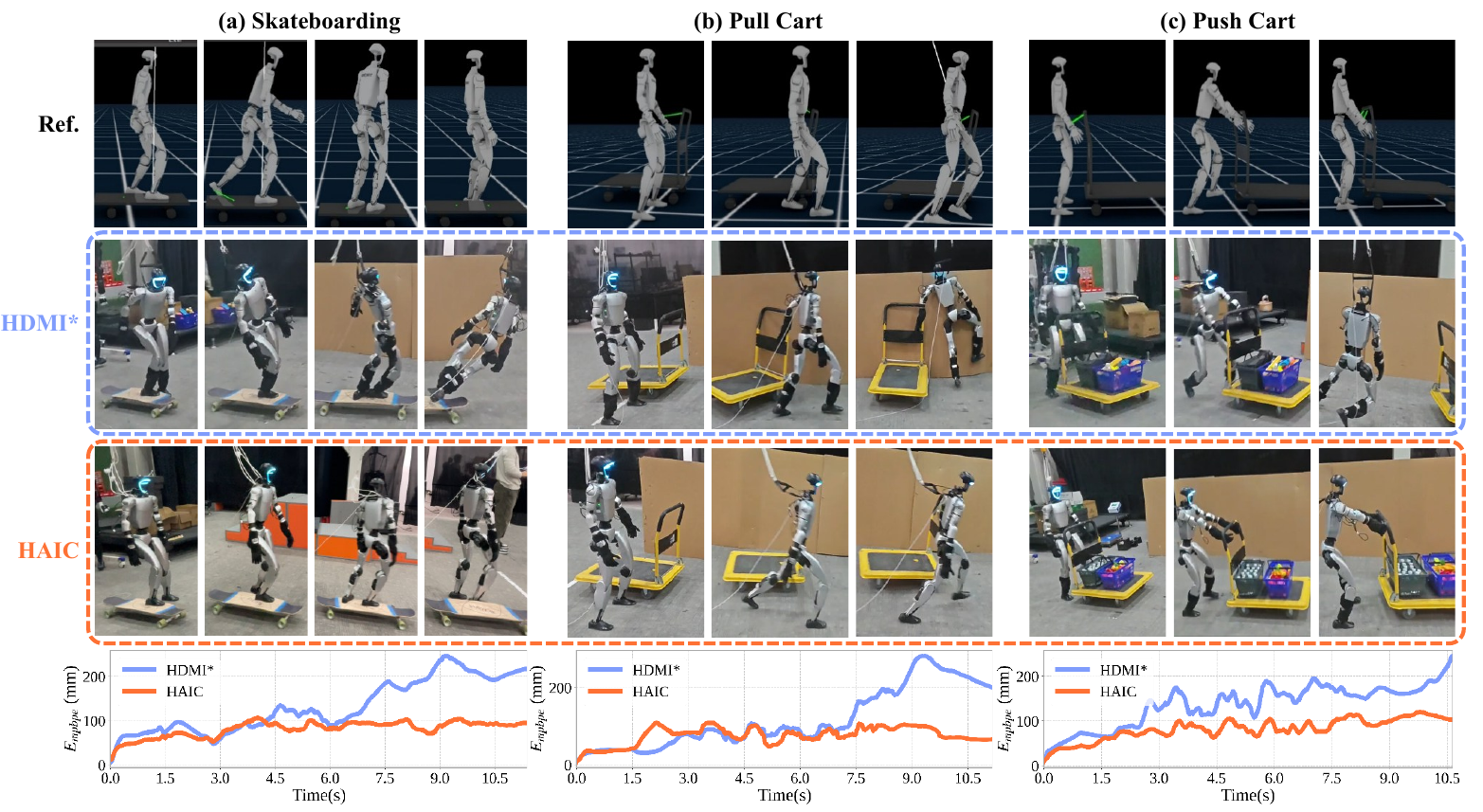}
    \caption{Real-world performance comparison with the baseline across various tasks. With the specifically designed dynamics-aware world model, \ours maintains robust stability throughout the interaction, whereas the baseline suffers from balance failures and trajectory drift.} 
    \label{fig:baseline_comparison}
    \vspace{-0.2in}
\end{figure*}

\subsection{Multiple Objects Contact Reward}
\label{sec:contact_reward}

To enable robust manipulation of multiple objects, we introduce a contact reward $r_{\text{contact}}$ that unifies geometric alignment and force regulation. As illustrated in Fig.~\ref{fig:contact_reward}, this reward dynamically guides the robot through two phases:
\begin{itemize}
    \item {Guidance (Fig.~\ref{fig:contact_reward}a):} Before contact, the reward minimizes the distance between end-effectors and semantic targets.
    \item {Execution} (Fig.~\ref{fig:contact_reward}b-c): Upon contact, it applies force appropriately while maintaining geometric constraints.
\end{itemize}

The contact reward is calculated as the average reward across all active objects $o \in \mathcal{O}$ and their corresponding end-effectors $e \in \mathcal{E}_o$:

\vspace{-0.2in} 
\begin{equation}
    r_{\text{contact}} = \frac{1}{|\mathcal{O}|} \sum_{o \in \mathcal{O}} \left( \frac{1}{|\mathcal{E}_o|} \sum_{e \in \mathcal{E}_o} \mathbb{I}_{o,e} \cdot r_{\text{pos}}^{o,e} \cdot r_{\text{force}}^{o,e} \right).
\end{equation}

Here, $\mathbb{I}_{o,e}$ is a binary mask from the reference motion indicating if contact is active. The individual terms account for physical tolerances to prevent optimization instability. Position term $r_{\text{pos}}^{o,e}$ encourages the end-effector position $\bm{p}_e$ to match the target $\bm{p}_{\text{tgt}}$ within a tolerance $\epsilon_{\text{tol}}$:
\begin{equation}
    r_{\text{pos}}^{o,e} = \exp \left( - {\max(0, \| \bm{p}_e - \bm{p}_{\text{tgt}} \| - \epsilon_{\text{tol}})}/{\sigma_p} \right).
\end{equation}
Force term $r_{\text{force}}^{o,e}$ ensures the contact force $\bm{F}_e$ reaches a required threshold $F_{\text{thr}}$. We penalize the force deficit only when it falls below this threshold:
\begin{equation}
    r_{\text{force}}^{o,e} = \exp \left( - {\max(0, F_{\text{thr}} - \| \bm{F}_e \|)}/{\sigma_f} \right).
\end{equation}
More detailed reward definitions and parameters are provided in Appendix~\ref{app:reward_details}.

%% file: sections/4_experiments.tex
\section{Experiments}
\label{sec:experiments}

\subsection{Experimenrt Setup}

\subsubsection{Task Description}
To validate our method, we designed three types of interaction tasks:

\noindent \textbf{Underactuated Object Interaction.} This task requires the robot to interact with underactuated wheeled objects, which introduces unstructured, time-varying external physical disturbances upon contact. Unlike interactions with fully actuated objects, the motion of underactuated objects actively affects the robot's balance, requiring the controller to predict and compensate for these disturbances in real time.

\noindent \textbf{Sequential interaction.} In addition to interacting with a single object, we also design composite sequential tasks involving multiple objects, such as loading a payload onto a mobile cart and then manipulating the cart. These scenarios challenge the policy's capacity to handle discrete mode switching and maintain stable control over extended horizons while coordinating interactions with multiple dynamic bodies.

\noindent \textbf{Multi-terrain interaction.} The final task requires the robot to carry a box across composite terrains, including platforms, slopes, and stairs. This task places high demands on state estimation because the box obscures the robot's view of the ground. As a result, the policy relies solely on self-perception and predictions from the world model. The controller must simultaneously handle the dynamic load on the arms and the complex interactions with the terrain.

\subsubsection{Implementation Details} We describe the specific configurations for data processing, training infrastructure, and deployment setups below.

\noindent \textbf{Data Collection and Processing.} To capture high-fidelity interaction behaviors, we utilized an optical motion capture system tracking both the human motion and interactive objects via synchronized markers. The raw kinematic data were retargeted to the Unitree G1 and refined through a physics-based simulation pipeline to generate physically valid states with contact information (see Appendix~\ref{app:dataset} for details).

\noindent \textbf{Training Setup.} We trained the base policy and world model using PPO in the Isaac Sim simulator. The two-stage training of a model was conducted on a single NVIDIA RTX 4090 GPU and took approximately 8 hours. 

\noindent \textbf{Sim-to-Real Transfer.} We employed extensive domain randomization to bridge the reality gap. For real-world deployment, the policy ran on an external PC, transmitting joint position targets to the robot's onboard control board via Ethernet. The PD controller operated at 200 Hz in simulation and 500 Hz on the real robot to ensure accurate tracking of the PD targets, while the control policies ran at 50 Hz.

\subsubsection{Evaluation Metrics}
To rigorously quantify the robot's kinematic tracking performance, we define the following metrics based on onboard proprioception. We report the Mean and Standard Deviation for error metrics, and the Success Rate ({SR}) for task completion.

\noindent \textbf{Robot State Metrics:}
\begin{itemize}[leftmargin=4mm]
    \item \textbf{$E_{\rm{mpbpe}}$ / $E_{\rm{mpboe}}$:} Root-relative Mean Per Body Position (mm) and Orientation ($10^{-3}$ rad) Error.
    \item \textbf{$E_{\rm{g\text{-}mpbpe}}$:} Global Mean Per Body Position Error (mm).
    \item \textbf{$E_{\rm{mpjpe}}$ / $E_{\rm{mpjve}}$:} Mean Per Joint Position ($10^{-3}$ rad) and Velocity ($10^{-3}$ rad/frame) Error.
    \item \textbf{$E_{\rm{mpbve}}$ / $E_{\rm{mpbae}}$:} Mean Per Body Velocity (mm/frame) and Acceleration (mm/frame$^2$) Error.
\end{itemize}

\noindent \textbf{Object State Metrics:}
\begin{itemize}[leftmargin=4mm]
    \item \textbf{$E_{\rm{mpope}}$ / $E_{\rm{mpooe}}$:} Mean Per Object Position (mm) and Orientation ($10^{-3}$ rad) Error.
    \item \textbf{$E_{\rm{mpove}}$ / $E_{\rm{mpoae}}$:} Mean Per Object Velocity (mm/frame) and Acceleration (mm/frame$^2$) Error.
\end{itemize}

We evaluate the performance of our proposed framework on the Unitree G1 humanoid robot in real-world scenarios. The robot operates autonomously using onboard computation, without external motion capture systems, relying solely on proprioceptive feedback and our dynamics-aware World Model.

\subsubsection{Baseline}
We compare our method with HDMI in simulation and the real world. We utilize \textbf{HDMI$^*$}, a proprioception-only variant of the original HDMI framework~\cite{weng2025hdmilearninginteractivehumanoid} as its lower bound. As illustrated in Fig.~\ref{fig:radar_comparison}, we also present the sim-to-sim upper bound performance of HDMI for comparison with the real-world success rates of HAIC and $\text{HDMI}^*$.
To ensure a fair comparison, this baseline was {retrained from scratch} using the same setup as our method. Instead of privileged object states, HDMI$^*$ receives a {history of proprioceptive observations} to implicitly infer the environmental context.

\subsection{Real-world Deployment and Comparison}

In real-world experiments, our dynamics-aware world model functions as a core estimator across three distinct scenarios:
(a) \textbf{Underactuated Object Interaction.} For underactuated objects like skateboards and carts, the world model infers the object's full dynamics, including translation, rotation, linear/angular velocity, and acceleration. These high-order predictions are explicitly applied to the object's canonical point cloud.
(b) \textbf{Sequential Interaction.} In sequential tasks involving multiple objects, \ours predicts the dynamics for both the box and the cart. It applies the respective predicted rotations and translations to align each object's geometric prior independently. 
(c) \textbf{Multi-terrain Interaction.} \ours extends its prediction to environmental features. It estimates the dynamics of the carried box while simultaneously predicting the translation and orientation of terrain structures. By projecting these estimates onto the geometry and feeding the aligned features into the downstream network, the robot successfully navigates uneven ground. We compared \ours with the baseline in these scenarios. \ours exhibits superior robustness due to its structured representation incorporating both geometric and dynamic information. The original HDMI performed well with fully-driven objects, but its performance declined on underactuated objects, since it does not account for geometry and higher-order physical quantities. As demonstrated by the case study results below, \ours shows significant advantages in both success rate and motion quality.

\input{tables/real_skateboard}
\input{tables/real_cart_manipulation}
\input{tables/real_task_2}

\input{tables/real_task_3}

\input{tables/sim_skateboard}
\input{tables/sim_cart_manipulation}

\subsubsection{Underactuated Object Interaction}
This task involves high-dynamic interactions, where object interactions significantly affect the robot's balance: skateboarding and manipulating wheeled carts. Fig.~\ref{fig:baseline_comparison} demonstrates that \ours maintains robust stability in skateboarding and cart tasks, whereas the baseline suffers from frequent failures and trajectory drift.
    
\noindent \textbf{Skateboarding:} Table~\ref{tab:real_skateboard} reports two success metrics: {glide success rate} (maintaining balance on the skateboard \textgreater 1s) and {complete success rate}. While the baseline struggles to maintain basic balance, \ours achieves a high glide success rate, demonstrating robust balance control. Although the complex dismount phase remains challenging, our method significantly outperforms the baseline, which fails to complete the task.

\noindent \textbf{Cart Manipulation:} Table~\ref{tab:real_carts} highlights the robustness of \ours. Notably, in the ``Push Cart'' task, the baseline fails as it cannot account for the cart's external dynamics and non-holonomic constraints. \ours predicts the object's forward velocity and acceleration, allowing the robot to proactively lean into the motion, resulting in a high success rate for both ``Pull'' and ``Push'' tasks.

\subsubsection{Sequential Interaction}
We evaluate sequential tasks requiring mode switching: ``Pull Cart w/ Box'' (load the box, then pull the cart) and ``Push Cart w/ Box'' (load the box, then push the cart). The baseline was unable to attempt these sequential composite tasks, as it cannot reliably complete a single cart manipulation task. Therefore, Table~\ref{tab:real_case3} reports only the performance of \ours. Despite the compounded complexity and error accumulation over a longer horizon, our framework enables the robot to transition smoothly, achieving a high success rate in the ``Push Cart w/ Box'' task.

\subsubsection{Multi-terrain Interaction}
We evaluate the robot's ability to carry a box across varying terrains, including flat ground, 15-degree slopes, stairs, and mixed slope-stair terrains. In these scenarios, \ours successfully picks and places a box across various types of terrains, where the policy completely occludes the robot's visual sensors. 
As shown in Table~\ref{tab:real_case1}, \ours demonstrates superior robustness on complex terrains. While the baseline handles simple slopes, it suffers catastrophic failures on the ``w/ Slope + Stair'' task due to error accumulation in state estimation. In contrast, \ours maintains a high success rate with lower orientation errors, validating that our geometric prior effectively compensates for the lack of visual feedback on uneven terrain.

\begin{figure*}[ht]
    \centering
    \includegraphics[width=\linewidth]{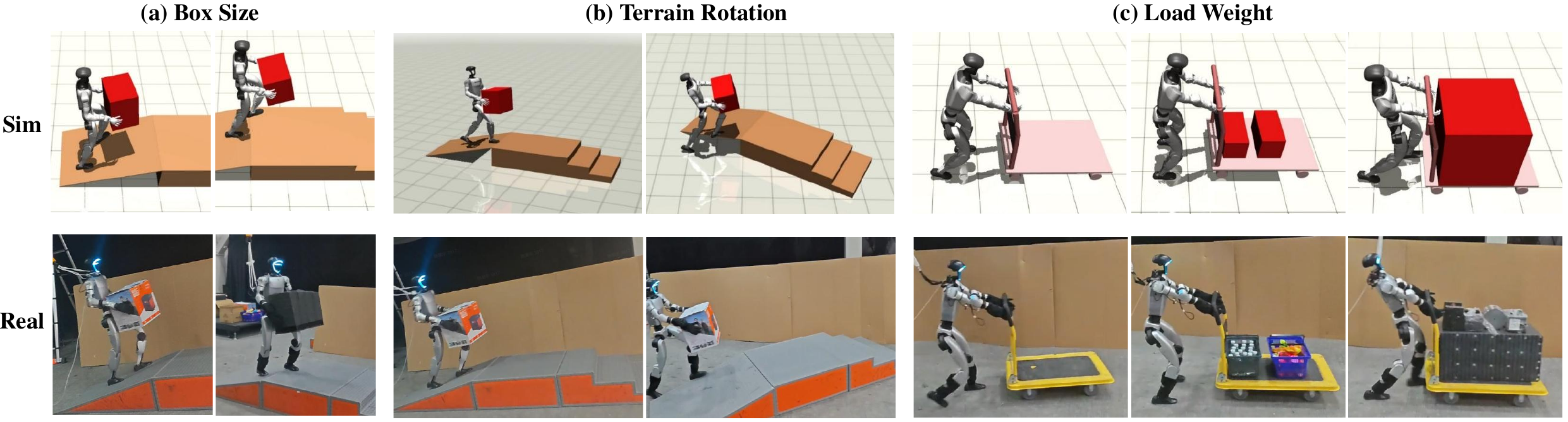}
    \caption{Sim-to-real performance across various HOI tasks, including pushing, carrying, and terrain traversal. \ours achieves robust interactions with diverse terrains and underactuated objects, and demonstrates strong generalization across object size, terrain orientation, and load weight.} 
    \label{fig:sim2real}
    \vspace{-0.2in}
\end{figure*}

\subsection{Ablation Study on \ours}

To rigorously analyze the contribution of each component in our framework—specifically, the benefits of higher-order dynamics prediction and explicit geometric projection—we define five model variants for our ablation study:

\begin{itemize}[leftmargin=4mm]
    \item \textbf{\ma:} A standard proprioception-only policy that observes only the robot's joint states and reference motion, without any explicit object representation.
    
    \item \textbf{\mb:} Extends the \ma by adding an auxiliary module that predicts the object's position and orientation ($\hat{p}, \hat{R}$). These are fed into the privilege adapter together.
    
    \item \textbf{\mc:} Extends Vec-Pose by predicting the comprehensive dynamics, including velocity and acceleration ($\hat{v}, \hat{a}$), which are also fed as flattened vector inputs.
    
    \item \textbf{\md:} Predicts the same state ($\hat{p}, \hat{R}$) as {Vec-Pose}. The predicted pose is used to transform the canonical point cloud into the robot frame. The explicit {geometric feature} and predicted pose are fed to the privilege adapter.
    
    \item \textbf{\ours:} The proposed full method. It predicts the comprehensive dynamics ($\hat{p}, \hat{R}, \hat{v}, \hat{a}$), projects the pose onto the geometric feature, and then feeds all the object's representations into the privilege adapter.
\end{itemize}

\begin{figure}[h]
    \centering
    \includegraphics[width=0.65\linewidth]{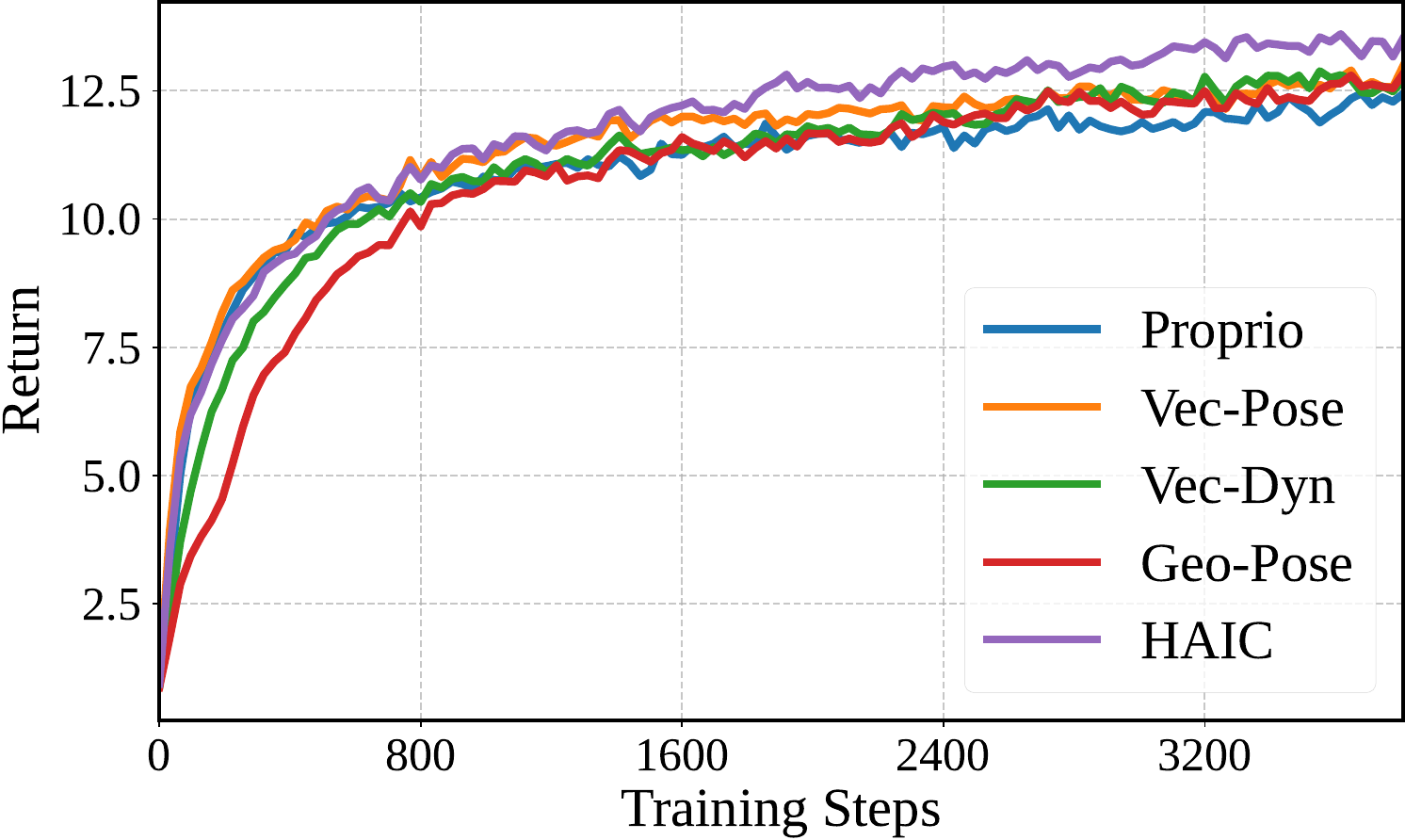}
    \caption{Learning curves on the ``Push Cart w/ Box'' task.} 
    \label{fig:abaltion_return}
    \vspace{-0.2in}
\end{figure}

\subsubsection{Analysis of Training Efficiency}
Fig.~\ref{fig:abaltion_return} illustrates the learning curves of all methods on the challenging {``Push Cart w/ Box''} task over 3,800 training steps. {\ours} demonstrates superior sample efficiency and achieves the highest asymptotic return compared to all baselines. We further evaluate the variants in the MuJoCo simulator within the same setup for quantitative comparison.

\subsubsection{Skateboarding} Table~\ref{tab:sim_skateboard} presents the results for the dynamic skateboarding task, where \ma struggles to maintain balance even during the glide phase, confirming the insufficiency of proprioception alone. Interestingly, while \mb achieves perfect gliding success, it fails during the dismount phase, indicating that position perception aids balance but lacks the foresight to handle impulsive forces upon stepping off. Comparing this with dynamics-aware methods highlights that predicting object acceleration is critical for managing transient disturbances. Consequently, \ours achieves high success rates across both phases with the lowest tracking errors, indicating that our world model generates necessary feedforward compensation for stability.

\subsubsection{Cart Manipulation} Table~\ref{tab:sim_carts_full} presents the results for cart manipulation. \ma fails to resolve object states, while \mb and \mc suffer from severe trajectory drift despite completing the task. In contrast, methods utilizing geometric projection (\md and \ours) achieve significantly tighter global tracking by strictly aligning with the cart's steering axis. Ultimately, \ours delivers the best overall performance by combining this geometric alignment with dynamics prediction for precise and stable manipulation.

\subsubsection{Other Complex Scenarios}
We further extend the evaluation to sequential and multi-terrain tasks. Detailed metrics are provided in Appendix Table~\ref{tab:sim_case2_full} and Table~\ref{tab:sim_case3_full}. Experimental results highlight two key advantages of {\ours}:
\begin{itemize}
    \item {Sequential Stability:} In sequential tasks like ``Pull Cart w/ Box'', baselines fail completely due to error accumulation. \ours leverages its world model to mitigate drift, thereby ensuring high success rates and precise object steering across extended interaction sequences.
    \item {Terrain Adaptability:} On the challenging ``Slope + Stair'' terrain, \ma and \mb suffer from stability issues. \ours maintains a high success rate by predicting dynamic disturbances caused by uneven ground.
\end{itemize}

\subsection{Robust interaction and Generalization}

To evaluate the policy's robustness and generalization capabilities when encountering unseen object attributes and environmental changes, we designed a series of rigorous variations, including object size, terrain rotation, and loaded weight. The results are summarized in Fig.~\ref{fig:sim2real}.

\subsubsection{Object Size and Weight Generalization} We tested beyond the training range ($0.1\text{m}$–$0.2\text{m}$, $1\text{kg}$–$2\text{kg}$). The input geometric prior $\mathcal{P}$ remained fixed as the original standard template. The policy successfully transported a large box with a side length of $0.25\text{m}$ and a heavy box weighing $3\text{kg}$. For heavy objects, the policy exhibited a lower center of gravity and a slower movement speed. For large-sized objects, the policy adjusted the grasping position and arm posture to avoid self-collisions. This indicates that the world model learned general relationships between mass, inertia, and motion dynamics.

\subsubsection{Terrain Orientation Perturbations} We randomly rotated slopes and stairs. The policy successfully adapted to these changes, maintaining stable walking by adjusting foot landing angles and body orientation. This demonstrates that the policy does not memorize specific terrain geometries but understands general equilibrium principles on inclined support surfaces.

\subsubsection{External Perturbations} We applied instantaneous lateral thrusts during robot carrying tasks. The strategy rapidly estimated changes in object and self-momentum caused by the disturbance using the world model, swiftly restoring balance through stepping and arm swinging, demonstrating the ability to adapt online to unknown perturbations.

\subsubsection{Extend experiment.} 
\begin{figure*}[ht]
    \centering
    \vspace{-0.1in}
    \includegraphics[width=1.0\linewidth]{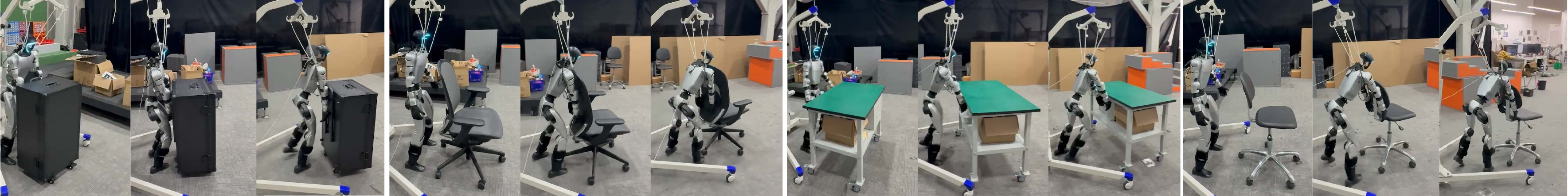}
    \caption{Real-world extend deployment experiment on everyday furniture. HAIC perform 100\% SR on table, suitcase and chairs.} 
    \label{fig:ExtendObjExp}
    \vspace{-0.1in}
\end{figure*}

We also extend the experiment to more underactuated interaction tasks on everyday furniture in Fig.~\ref{fig:ExtendObjExp}, and HAIC achieves a 100\% success rate on all tasks, including pushing a table, a suitcase, and chairs with an extra yaw freedom on the shaft.

These experiments demonstrate that by combining geometric priors and predictions of dynamics in end-to-end training, \ours can acquire an understanding of the essence of physical interactions, and exhibit zero-shot generalization capabilities.
%

\subsection{Analysis and Explanation of world model}
\subsubsection{Comparison on world model}
To evaluate the effectiveness of our explicit world modeling approach, we compare \ours against Rapid Motor Adaptation (RMA)~\cite{kumar2021rma}. While RMA relies on an implicit environmental latent to adapt to dynamics, our DWM explicitly models the object's physical behaviors. As summarized in Table \ref{tab:rma_compare}, \ours achieves superior or competitive success rates and significantly lower tracking errors across most kinematic metrics in both cart manipulation tasks. These results validate that for handling complex underactuated interactions, explicit dynamics prediction is more robust and effective than implicit adaptation.
\input{tables/rma_comparison}

 \begin{figure}
     \centering
     \includegraphics[width=\linewidth]{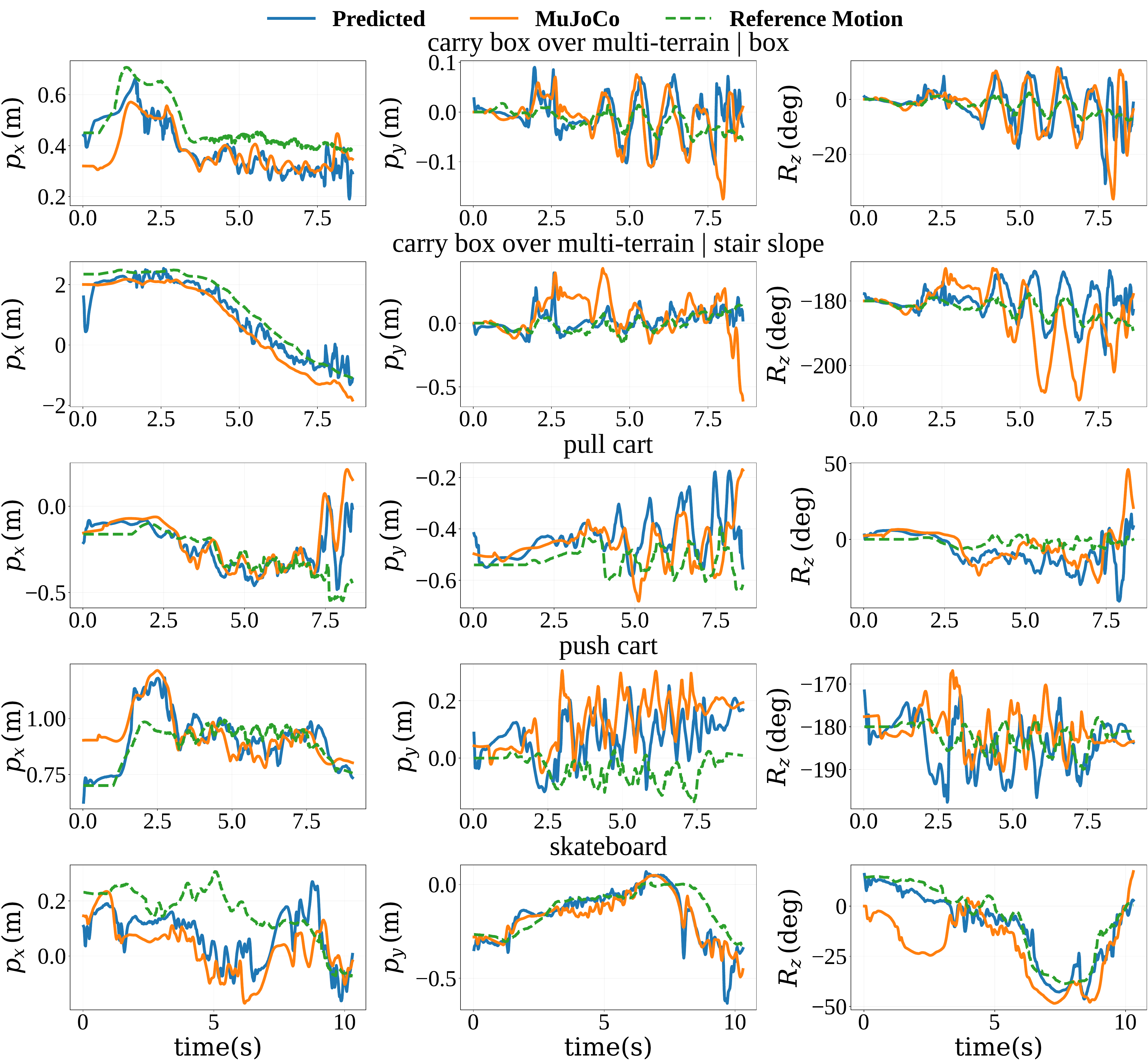}
     \caption{Comparison of the relative object states in robot's local frame from dynamics-aware world model, MuJoCo and reference motion.}
     \label{fig:analysis_dwm}
     \vspace{-0.2in}
 \end{figure}

 \subsubsection{DWM and Failure case analysis} We visualize the robot frame object state from DWM prediction, MuJoCo, and reference motion in Fig.~\ref{fig:analysis_dwm}. We select several highly salient object state components, namely translation along x or y-axis, and rotation on z-axis. For box state, the prediction closely match the simulated results. However, for challenging tasks like pushing or pulling cart and skateboarding, the predicted states exhibit drift from the simulated trajectories, tending to revert toward the reference motion state. This error eventually leads to an observation mismatch, preventing it from adapting effectively. When its prediction of the object state converges within a error threshold, it can ensure the policy success. When skateboard decelerates and lands, divergence from the actual state will lead to failure (60\% SR). As illustrated in Fig.~\ref{fig:analysis_dwm}, transforming the static terrain into the robot's local frame equips the DWM with crucial spatial awareness. This enables the system to actively alleviate heading drifts and maintain the intended path, even under terrain rotations.
 
\subsubsection{Robustness to Hyperparameters and Annotations}  
As shown in Table~\ref{tab:ablation_thresholds}, even with imprecise thresholds, HAIC maintains a high success rate with only minor kinematic degradation. Furthermore, our binary contact masks are automatically computed via a standardized script rather than manually annotated. This algorithmic approach is generalizable across tasks and avoids human annotation noise.
\input{tables/sensitivity_analysis}

%% file: tables/real_skateboard.tex

\begin{table}[h]
\centering
\colorlet{lightorange}{orange!25}

\caption{Real-world results for \textbf{Skateboarding}. We report the success rate for both Gliding and Completing. \ours achieves stable gliding and significantly higher completion rates compared to the baseline.}
\label{tab:real_skateboard}
\resizebox{\columnwidth}{!}{
\begin{tabular}{lcccccccc}
\toprule
& \multicolumn{2}{c}{\textbf{Success Rate}} & \multicolumn{6}{c}{\textbf{Robot State Metrics}} \\
\textbf{Method} & \textbf{Glide} & \textbf{Comp.} & \textbf{$E_\mathrm{mpbpe}$} & \textbf{$E_\mathrm{mpboe}$} & \textbf{$E_\mathrm{mpjpe}$} & \textbf{$E_\mathrm{mpbve}$} & \textbf{$E_\mathrm{mpbae}$} & \textbf{$E_\mathrm{mpjve}$} \\
\midrule
\baseline & 20\% & 0\% & 132.1$^{\pm 35.7}$ & 666.7$^{\pm 213.6}$ & 188.7$^{\pm 12.6}$ & 5.39$^{\pm 0.62}$ & \cellcolor{lightorange}\textbf{4.06$^{\pm 0.04}$} & \cellcolor{lightorange}\textbf{10.40$^{\pm 0.19}$} \\
\textbf{\ours} & \cellcolor{lightorange}\textbf{100\%} & \cellcolor{lightorange}\textbf{60\%} & \cellcolor{lightorange}\textbf{81.5$^{\pm 15.4}$} & \cellcolor{lightorange}\textbf{453.5$^{\pm 88.2}$} & \cellcolor{lightorange}\textbf{127.1$^{\pm 11.5}$} & \cellcolor{lightorange}\textbf{4.72$^{\pm 0.75}$} & 4.48$^{\pm 0.50}$ & 12.15$^{\pm 1.87}$ \\
\bottomrule
\end{tabular}}
\vspace{-10pt} 
\end{table}

%% file: tables/real_cart_manipulation.tex

\begin{table}[h]
\centering
\colorlet{lightorange}{orange!25}

\caption{Real-world results for \textbf{Cart Manipulation}. \ours achieves 100\% success in both tasks, whereas the baseline fails in ``Push''.}
\label{tab:real_carts}
\resizebox{\columnwidth}{!}{
\begin{tabular}{llccccccc}
\toprule
\textbf{Task} & \textbf{Method} & \textbf{SR} $\uparrow$ & \textbf{$E_\mathrm{mpbpe}$} $\downarrow$ & \textbf{$E_\mathrm{mpboe}$} $\downarrow$ & \textbf{$E_\mathrm{mpjpe}$} $\downarrow$ & \textbf{$E_\mathrm{mpbve}$} $\downarrow$ & \textbf{$E_\mathrm{mpbae}$} $\downarrow$ & \textbf{$E_\mathrm{mpjve}$} $\downarrow$ \\
\midrule
\multirow{2}{*}{\makecell[l]{Pull\\Cart}} 
& \baseline & 40\% & 118.0$^{\pm 5.8}$ & 556.7$^{\pm 43.0}$ & 165.0$^{\pm 2.3}$ & 6.11$^{\pm 0.13}$ & 3.73$^{\pm 0.14}$ & 10.32$^{\pm 1.00}$ \\
& \textbf{\ours} & \cellcolor{lightorange}\textbf{100\%} & \cellcolor{lightorange}\textbf{76.2$^{\pm 6.8}$} & \cellcolor{lightorange}\textbf{425.0$^{\pm 41.7}$} & \cellcolor{lightorange}\textbf{132.3$^{\pm 6.4}$} & \cellcolor{lightorange}\textbf{5.08$^{\pm 0.53}$} & \cellcolor{lightorange}\textbf{3.66$^{\pm 0.21}$} & \cellcolor{lightorange}\textbf{9.50$^{\pm 1.10}$} \\
\midrule
\multirow{2}{*}{\makecell[l]{Push\\Cart}} 
& \baseline & 0\% & 136.2$^{\pm 23.3}$ & 778.5$^{\pm 218.6}$ & {158.9$^{\pm 4.7}$} & 7.36$^{\pm 0.48}$ & 5.32$^{\pm 0.21}$ & 15.62$^{\pm 0.74}$ \\
& \textbf{\ours} & \cellcolor{lightorange}\textbf{100\%} & \cellcolor{lightorange}\textbf{82.0$^{\pm 6.3}$} & \cellcolor{lightorange}\textbf{418.7$^{\pm 36.6}$} & \cellcolor{lightorange}\textbf{150.0}$^{\pm 9.3}$ & \cellcolor{lightorange}\textbf{5.29$^{\pm 0.17}$} & \cellcolor{lightorange}\textbf{4.47$^{\pm 0.09}$} & \cellcolor{lightorange}\textbf{11.61$^{\pm 0.80}$} \\
\bottomrule
\end{tabular}}
\vspace{-10pt} 
\end{table}

%% file: tables/real_task_2.tex

\begin{table}[ht]
\centering
\colorlet{lightorange}{orange!25}

\caption{Real-world results for \textbf{Sequential Interaction}. The baseline is omitted as it failed to complete the prerequisite cart tasks. \ours demonstrates capability in sequential object transport.}
\label{tab:real_case3}
\resizebox{\columnwidth}{!}{
\begin{tabular}{llccccccc}
\toprule
\textbf{Task} & \textbf{Method} & \textbf{SR} $\uparrow$ & \textbf{$E_\mathrm{mpbpe}$} $\downarrow$ & \textbf{$E_\mathrm{mpboe}$} $\downarrow$ & \textbf{$E_\mathrm{mpjpe}$} $\downarrow$ & \textbf{$E_\mathrm{mpbve}$} $\downarrow$ & \textbf{$E_\mathrm{mpbae}$} $\downarrow$ & \textbf{$E_\mathrm{mpjve}$} $\downarrow$ \\
\midrule
\multirow{2}{*}{\makecell[l]{Pull Cart\\ w/ Box}} & \baseline & 0\% & - & - & - & - & - & - \\
& \textbf{\ours} & \cellcolor{lightorange}\textbf{40\%} & \cellcolor{lightorange}\textbf{88.2$^{\pm 8.2}$} & \cellcolor{lightorange}\textbf{420.7$^{\pm 16.7}$} & \cellcolor{lightorange}\textbf{157.2$^{\pm 8.2}$} & \cellcolor{lightorange}\textbf{5.91$^{\pm 0.07}$} & \cellcolor{lightorange}\textbf{5.02$^{\pm 0.09}$} & \cellcolor{lightorange}\textbf{12.20$^{\pm 0.31}$} \\
\midrule
\multirow{2}{*}{\makecell[l]{Push Cart\\ w/ Box}} & \baseline & 0\% & - & - & - & - & - & - \\
& \textbf{\ours} & \cellcolor{lightorange}\textbf{100\%} & \cellcolor{lightorange}\textbf{70.5$^{\pm 5.7}$} & \cellcolor{lightorange}\textbf{381.7$^{\pm 24.7}$} & \cellcolor{lightorange}\textbf{143.6$^{\pm 4.1}$} & \cellcolor{lightorange}\textbf{5.27$^{\pm 0.26}$} & \cellcolor{lightorange}\textbf{3.92$^{\pm 0.18}$} & \cellcolor{lightorange}\textbf{11.10$^{\pm 0.85}$} \\
\bottomrule
\end{tabular}}
\vspace{-10pt}
\end{table}

%% file: tables/real_task_3.tex
\begin{table}[h]
\centering
\colorlet{lightorange}{orange!25}

\caption{Real-world results for \textbf{Multi-terrain Interaction}. \ours succeeds on the complex slope-stair terrain where baseline fails.}
\label{tab:real_case1}
\resizebox{\columnwidth}{!}{
\begin{tabular}{llccccccc}
\toprule
\textbf{Task} & \textbf{Method} & \textbf{SR} $\uparrow$ & \textbf{$E_\mathrm{mpbpe}$} $\downarrow$ & \textbf{$E_\mathrm{mpboe}$} $\downarrow$ & \textbf{$E_\mathrm{mpjpe}$} $\downarrow$ & \textbf{$E_\mathrm{mpbve}$} $\downarrow$ & \textbf{$E_\mathrm{mpbae}$} $\downarrow$ & \textbf{$E_\mathrm{mpjve}$} $\downarrow$ \\
\midrule
\multirow{2}{*}{\makecell[l]{Carry\\Box}} 
& \baseline & 100\% & 66.6$^{\pm 2.0}$ & 344.0$^{\pm 19.8}$ & 171.7$^{\pm 1.7}$ & 4.69$^{\pm 0.10}$ & 3.14$^{\pm 0.01}$ & 8.10$^{\pm 0.13}$ \\
& \textbf{\ours} & 100\% & \cellcolor{lightorange}\textbf{58.8$^{\pm 6.1}$} & \cellcolor{lightorange}\textbf{305.6$^{\pm 20.8}$} & \cellcolor{lightorange}\textbf{162.0$^{\pm 6.8}$} & \cellcolor{lightorange}\textbf{4.21$^{\pm 0.13}$} & \cellcolor{lightorange}\textbf{3.03$^{\pm 0.05}$} & \cellcolor{lightorange}\textbf{7.75$^{\pm 0.18}$} \\
\midrule
\multirow{2}{*}{{w/ Stair}} 
& \baseline & 100\% & 66.2$^{\pm 4.1}$ & \cellcolor{lightorange}\textbf{316.2$^{\pm 2.8}$} & 158.9$^{\pm 13.8}$ & 4.79$^{\pm 0.36}$ & 4.35$^{\pm 0.24}$ & 10.10$^{\pm 0.98}$ \\
& \textbf{\ours} & 100\% & \cellcolor{lightorange}\textbf{55.5$^{\pm 3.9}$} & 331.2$^{\pm 23.1}$ & \cellcolor{lightorange}\textbf{118.4$^{\pm 0.6}$} & \cellcolor{lightorange}\textbf{4.06$^{\pm 0.02}$} & \cellcolor{lightorange}\textbf{4.05$^{\pm 0.01}$} & \cellcolor{lightorange}\textbf{8.66$^{\pm 0.01}$} \\
\midrule
\multirow{2}{*}{{w/ Slope}} 
& \baseline & 100\% & \cellcolor{lightorange}\textbf{53.5$^{\pm 0.7}$} & \cellcolor{lightorange}\textbf{289.0$^{\pm 1.7}$} & 130.6$^{\pm 1.9}$ & \cellcolor{lightorange}\textbf{3.68$^{\pm 0.03}$} & \cellcolor{lightorange}\textbf{3.39$^{\pm 0.01}$} & 7.41$^{\pm 0.02}$ \\
& \textbf{\ours} & 100\% & 60.5$^{\pm 4.3}$ & 324.7$^{\pm 28.0}$ & \cellcolor{lightorange}\textbf{120.2$^{\pm 1.6}$} & 8.77$^{\pm 0.11}$ & 4.60$^{\pm 0.13}$ & \cellcolor{lightorange}\textbf{3.66$^{\pm 0.05}$} \\
\midrule
\multirow{2}{*}{\makecell[l]{w/ Stair\\+ Slope}} 
& \baseline & 0\% & 126.8$^{\pm 4.4}$ & 701.0$^{\pm 32.1}$ & 162.3$^{\pm 2.0}$ & 5.98$^{\pm 0.07}$ & \cellcolor{lightorange}\textbf{4.03$^{\pm 0.10}$} & 10.87$^{\pm 0.27}$ \\
& \textbf{\ours} & \cellcolor{lightorange}\textbf{100\%} & \cellcolor{lightorange}\textbf{60.6$^{\pm 13.4}$} & \cellcolor{lightorange}\textbf{286.5$^{\pm 59.8}$} & \cellcolor{lightorange}\textbf{128.5$^{\pm 16.2}$} & \cellcolor{lightorange}\textbf{4.74$^{\pm 0.17}$} & 4.05$^{\pm 0.04}$ & \cellcolor{lightorange}\textbf{10.67$^{\pm 0.52}$} \\
\bottomrule
\end{tabular}}
\vspace{-10pt}
\end{table}

%% file: tables/sim_skateboard.tex
\begin{table*}[ht]
\centering
\colorlet{lightorange}{orange!25}
\colorlet{lighterorange}{orange!12}

\caption{Ablation results for \textbf{Skateboarding}. {\mb} can balance but fails to dismount due to a lack of acceleration prediction. {\ours} achieves a high completion rate with the lowest robot tracking errors and the most stable object acceleration control.}
\label{tab:sim_skateboard}
\resizebox{\textwidth}{!}{
\begin{tabular}{lcclllllllllll}
\toprule
& \multicolumn{2}{c}{\textbf{SR} $\uparrow$} & \multicolumn{7}{c}{\textbf{Robot State Metrics}} & \multicolumn{4}{c}{\textbf{Object State Metrics}} \\
\cmidrule(lr){2-3} \cmidrule(lr){4-10} \cmidrule(lr){11-14}
\textbf{Method} & \textbf{Glide} & \textbf{Comp.} & \textbf{$E_\mathrm{mpbpe}$}$\downarrow$ & \textbf{$E_\mathrm{g\text{-}mpbpe}$}$\downarrow$ & \textbf{$E_\mathrm{mpboe}$}$\downarrow$ & \textbf{$E_\mathrm{mpjpe}$}$\downarrow$ & \textbf{$E_\mathrm{mpbve}$}$\downarrow$ & \textbf{$E_\mathrm{mpbae}$}$\downarrow$ & \textbf{$E_\mathrm{mpjve}$}$\downarrow$ & \textbf{$E_\mathrm{mpope}$}$\downarrow$ & \textbf{$E_\mathrm{mpooe}$}$\downarrow$ & \textbf{$E_\mathrm{mpove}$}$\downarrow$ & \textbf{$E_\mathrm{mpoae}$}$\downarrow$ \\
\midrule
\ma & 40\% & 0\% & 91.1$^{\pm 9.6}$ & 548.1$^{\pm 186.4}$ & 490.6$^{\pm 58.0}$ & 163.3$^{\pm 3.4}$ & 5.08$^{\pm 0.28}$ & 4.18$^{\pm 0.10}$ & 12.45$^{\pm 0.55}$ & 549.9$^{\pm 167.5}$ & 1219$^{\pm 179}$ & 9.63$^{\pm 0.64}$ & 11.02$^{\pm 0.25}$ \\
\mb & 100\% & 0\% & 77.1$^{\pm 12.7}$ & \cellcolor{lighterorange}430.3$^{\pm 248.2}$ & 394.0$^{\pm 55.0}$ & 167.0$^{\pm 4.7}$ & \cellcolor{lightorange}\textbf{4.25$^{\pm 0.20}$} & \cellcolor{lightorange}\textbf{3.94$^{\pm 0.09}$} & \cellcolor{lightorange}\textbf{10.14$^{\pm 0.10}$} & \cellcolor{lighterorange}434.8$^{\pm 259.9}$ & 1192$^{\pm 147}$ & 8.78$^{\pm 1.25}$ & 10.75$^{\pm 0.10}$ \\
\mc & 100\% & 100\% & 74.6$^{\pm 0.8}$ & 444.7$^{\pm 13.3}$ & 416.7$^{\pm 6.8}$ & 147.7$^{\pm 2.3}$ & 4.87$^{\pm 0.16}$ & 4.11$^{\pm 0.16}$ & \cellcolor{lighterorange}12.20$^{\pm 0.56}$ & \cellcolor{lightorange}\textbf{432.1$^{\pm 9.1}$} & \cellcolor{lighterorange}1038$^{\pm 137}$ & \cellcolor{lighterorange}8.64$^{\pm 0.36}$ & 10.90$^{\pm 0.28}$ \\
\md & 100\% & 100\% & \cellcolor{lighterorange}60.2$^{\pm 2.5}$ & 513.0$^{\pm 86.2}$ & \cellcolor{lighterorange}354.5$^{\pm 18.3}$ & \cellcolor{lighterorange}121.6$^{\pm 1.8}$ & 4.55$^{\pm 0.08}$ & \cellcolor{lighterorange}4.01$^{\pm 0.04}$ & 12.42$^{\pm 0.22}$ & 565.8$^{\pm 83.2}$ & 1071$^{\pm 72}$ & 8.77$^{\pm 0.98}$ & \cellcolor{lighterorange}10.74$^{\pm 0.11}$ \\
\textbf{\ours} & 100\% & 100\% & \cellcolor{lightorange}\textbf{57.8$^{\pm 2.8}$} & \cellcolor{lightorange}\textbf{417.8$^{\pm 36.4}$} & \cellcolor{lightorange}\textbf{335.5$^{\pm 19.7}$} & \cellcolor{lightorange}\textbf{117.6$^{\pm 0.7}$} & \cellcolor{lighterorange}4.54$^{\pm 0.11}$ & 4.04$^{\pm 0.05}$ & 12.30$^{\pm 0.31}$ & 473.8$^{\pm 43.1}$ & \cellcolor{lightorange}\textbf{1035$^{\pm 57}$} & \cellcolor{lightorange}\textbf{8.08$^{\pm 0.32}$} & \cellcolor{lightorange}\textbf{10.66$^{\pm 0.01}$} \\
\bottomrule
\end{tabular}}
\end{table*}

%% file: tables/sim_cart_manipulation.tex
\begin{table*}[ht]
\centering
\colorlet{lightorange}{orange!25}   
\colorlet{lighterorange}{orange!12} 

\caption{Ablation results for \textbf{Cart Manipulation}. {\ours} shows consistently high performance, achieving either the best or second-best results across most metrics.}
\label{tab:sim_carts_full}
\resizebox{\textwidth}{!}{
\begin{tabular}{llclllllllllll}
\toprule
& & & \multicolumn{7}{c}{\textbf{Robot State Metrics}} & \multicolumn{4}{c}{\textbf{Object State Metrics}} \\
\cmidrule(lr){4-10} \cmidrule(lr){11-14}
\textbf{Task} & \textbf{Method} & \textbf{SR} $\uparrow$ & \textbf{$E_\mathrm{mpbpe}$}$\downarrow$ & \textbf{$E_\mathrm{g\text{-}mpbpe}$}$\downarrow$ & \textbf{$E_\mathrm{mpboe}$}$\downarrow$ & \textbf{$E_\mathrm{mpjpe}$}$\downarrow$ & \textbf{$E_\mathrm{mpbve}$}$\downarrow$ & \textbf{$E_\mathrm{mpbae}$}$\downarrow$ & \textbf{$E_\mathrm{mpjve}$}$\downarrow$ & \textbf{$E_\mathrm{mpope}$}$\downarrow$ & \textbf{$E_\mathrm{mpooe}$}$\downarrow$ & \textbf{$E_\mathrm{mpove}$}$\downarrow$ & \textbf{$E_\mathrm{mpoae}$}$\downarrow$ \\
\midrule

\multirow{5}{*}{\makecell[l]{Pull\\Cart}} 
& \ma & 60\% & 97.9$^{\pm 34.7}$ & 341.9$^{\pm 114.6}$ & 575.4$^{\pm 188}$ & 137.4$^{\pm 6.5}$ & 5.58$^{\pm 0.73}$ & 3.75$^{\pm 0.22}$ & 11.39$^{\pm 1.69}$ & 448.8$^{\pm 190}$ & 1551$^{\pm 149}$ & 5.90$^{\pm 2.41}$ & 3.59$^{\pm 0.06}$ \\
& \mb & 60\% & 100.0$^{\pm 26.6}$ & 388.0$^{\pm 221.0}$ & 589.6$^{\pm 141}$ & 145.3$^{\pm 13.5}$ & 5.86$^{\pm 0.74}$ & 3.83$^{\pm 0.31}$ & 11.96$^{\pm 2.06}$ & 437.7$^{\pm 223}$ & 1530$^{\pm 212}$ & 5.78$^{\pm 1.96}$ & \cellcolor{lighterorange}3.55$^{\pm 0.03}$ \\
& \mc & 100\% & 71.3$^{\pm 0.5}$ & \cellcolor{lighterorange}244.1$^{\pm 22.3}$ & 406.9$^{\pm 9.3}$ & 133.7$^{\pm 1.0}$ & \cellcolor{lighterorange}5.12$^{\pm 0.04}$ & \cellcolor{lighterorange}3.54$^{\pm 0.01}$ & \cellcolor{lighterorange}9.88$^{\pm 0.18}$ & \cellcolor{lighterorange}230.1$^{\pm 16.0}$ & 1640$^{\pm 108}$ & \cellcolor{lightorange}\textbf{4.05$^{\pm 0.02}$} & \cellcolor{lightorange}\textbf{3.52$^{\pm 0.01}$} \\
& \md & 100\% & \cellcolor{lighterorange}66.1$^{\pm 0.3}$ & 261.5$^{\pm 32.7}$ & \cellcolor{lighterorange}384.1$^{\pm 21.7}$ & \cellcolor{lighterorange}127.3$^{\pm 4.4}$ & 5.29$^{\pm 0.66}$ & 3.65$^{\pm 0.19}$ & 10.85$^{\pm 1.52}$ & 320.3$^{\pm 89.5}$ & \cellcolor{lightorange}\textbf{1210$^{\pm 11.6}$} & \cellcolor{lighterorange}4.09$^{\pm 0.11}$ & 3.57$^{\pm 0.04}$ \\
& \textbf{\ours} & 100\% & \cellcolor{lightorange}\textbf{64.8$^{\pm 2.8}$} & \cellcolor{lightorange}\textbf{225.7$^{\pm 3.4}$} & \cellcolor{lightorange}\textbf{372.7$^{\pm 34.1}$} & \cellcolor{lightorange}\textbf{123.1$^{\pm 0.3}$} & \cellcolor{lightorange}\textbf{4.75$^{\pm 0.01}$} & \cellcolor{lightorange}\textbf{3.48$^{\pm 0.01}$} & \cellcolor{lightorange}\textbf{9.39$^{\pm 0.09}$} & \cellcolor{lightorange}\textbf{219.5$^{\pm 6.6}$} & \cellcolor{lighterorange}1241$^{\pm 247}$ & 4.19$^{\pm 0.02}$ & \cellcolor{lighterorange}3.55$^{\pm 0.01}$ \\
\midrule

\multirow{5}{*}{\makecell[l]{Push\\Cart}} 
& \ma & 0\% & 112.9$^{\pm 9.2}$ & 276.6$^{\pm 80.5}$ & 538.4$^{\pm 129.2}$ & 172.1$^{\pm 14.2}$ & 6.50$^{\pm 0.41}$ & 4.78$^{\pm 0.14}$ & 15.38$^{\pm 0.85}$ & 460.7$^{\pm 286.2}$ & 1406$^{\pm 16}$ & 4.61$^{\pm 0.91}$ & 3.75$^{\pm 0.04}$ \\
& \mb & 100\% & \cellcolor{lighterorange}61.5$^{\pm 1.8}$ & 451.8$^{\pm 82.9}$ & \cellcolor{lightorange}\textbf{287.7$^{\pm 3.1}$} & \cellcolor{lightorange}\textbf{125.3$^{\pm 2.0}$} & \cellcolor{lighterorange}4.65$^{\pm 0.17}$ & \cellcolor{lighterorange}4.16$^{\pm 0.10}$ & \cellcolor{lighterorange}11.20$^{\pm 0.91}$ & 338.7$^{\pm 63.9}$ & \cellcolor{lightorange}\textbf{1152$^{\pm 58}$} & 4.25$^{\pm 0.43}$ & 3.71$^{\pm 0.01}$ \\
& \mc & 100\% & 62.3$^{\pm 2.5}$ & 439.0$^{\pm 75.7}$ & \cellcolor{lighterorange}299.9$^{\pm 16.3}$ & \cellcolor{lighterorange}126.4$^{\pm 2.2}$ & \cellcolor{lightorange}\textbf{4.58$^{\pm 0.05}$} & \cellcolor{lightorange}\textbf{4.12$^{\pm 0.04}$} & \cellcolor{lightorange}\textbf{10.86$^{\pm 0.39}$} & 358.4$^{\pm 73.1}$ & 1172$^{\pm 26}$ & 4.63$^{\pm 1.03}$ & \cellcolor{lighterorange}\textbf{3.70$^{\pm 0.01}$} \\
& \md & 100\% & 64.0$^{\pm 4.8}$ & \cellcolor{lighterorange}249.1$^{\pm 43.3}$ & 346.8$^{\pm 28.5}$ & 141.0$^{\pm 3.1}$ & 4.93$^{\pm 0.06}$ & 4.28$^{\pm 0.03}$ & 12.70$^{\pm 0.16}$ & \cellcolor{lightorange}\textbf{259.8$^{\pm 66.8}$} & \cellcolor{lighterorange}1153$^{\pm 107}$ & \cellcolor{lightorange}\textbf{3.65$^{\pm 0.41}$} & \cellcolor{lighterorange}\textbf{3.70$^{\pm 0.01}$} \\
& \textbf{\ours} & 100\% & \cellcolor{lightorange}\textbf{61.3$^{\pm 2.1}$} & \cellcolor{lightorange}\textbf{221.0$^{\pm 74.3}$} & 319.9$^{\pm 13.0}$ & 145.9$^{\pm 5.5}$ & 5.00$^{\pm 0.10}$ & 4.23$^{\pm 0.08}$ & 12.36$^{\pm 0.39}$ & \cellcolor{lighterorange}267.9$^{\pm 85.6}$ & 1262$^{\pm 96}$ & \cellcolor{lighterorange}4.04$^{\pm 1.07}$ & \cellcolor{lightorange}\textbf{3.70$^{\pm 0.01}$} \\
\bottomrule
\end{tabular}}
\end{table*}

%% file: tables/rma_comparison.tex
\begin{table}[ht]
\centering
\colorlet{lightorange}{orange!25}

\caption{Compared to the RMA baseline, \ours achieves superior or competitive success rates and significantly lower tracking errors.}
\label{tab:rma_compare}
\resizebox{\columnwidth}{!}{
\begin{tabular}{llccccccc}
\toprule
\textbf{Task} & \textbf{Method} & \textbf{SR} $\uparrow$ & \textbf{$E_\mathrm{mpbpe}$} $\downarrow$ & \textbf{$E_\mathrm{g\text{-}mpbpe}$}$\downarrow$& \textbf{$E_\mathrm{mpboe}$} $\downarrow$ & \textbf{$E_\mathrm{mpjpe}$} $\downarrow$ & \textbf{$E_\mathrm{mpope}$} $\downarrow$ & \textbf{$E_\mathrm{mpooe}$} $\downarrow$ \\
\midrule
\multirow{2}{*}{\makecell[l]{Pull Cart}} & RMA & 70\% & 99.1$^{\pm 38.2}$ & 323.1$^{\pm 110.8}$ & 530.6$^{\pm 168.6}$ & 139.9$^{\pm 5.6}$ & 530.6$^{\pm 168.6}$ & 1474$^{\pm 178.5}$ \\
& \textbf{\ours} & \cellcolor{lightorange}\textbf{100\%} & \cellcolor{lightorange}\textbf{64.8$^{\pm 2.8}$} & \cellcolor{lightorange}\textbf{225.7$^{\pm 3.4}$} & \cellcolor{lightorange}\textbf{372.7$^{\pm 34.1}$} & \cellcolor{lightorange}\textbf{123.1$^{\pm 0.3}$} & \cellcolor{lightorange}\textbf{219.5$^{\pm 6.6}$} & \cellcolor{lightorange}\textbf{1241$^{\pm 246.5}$} \\
\midrule
\multirow{2}{*}{\makecell[l]{Push Cart}} & RMA & \cellcolor{lightorange}\textbf{100\%} & 74.1$^{\pm 6.3}$ & 267.0$^{\pm 52.8}$ & 340.8$^{\pm 30.2}$ & \cellcolor{lightorange}\textbf{140.1$^{\pm 6.0}$} & 519.9$^{\pm 92.9}$ & \cellcolor{lightorange}\textbf{1120$^{\pm 31.0}$} \\
& \textbf{\ours} & \cellcolor{lightorange}\textbf{100\%} & \cellcolor{lightorange}\textbf{61.3$^{\pm 2.1}$} & \cellcolor{lightorange}\textbf{221.0$^{\pm 74.3}$} & \cellcolor{lightorange}\textbf{319.9$^{\pm 13.0}$} & 145.9$^{\pm 5.5}$ & \cellcolor{lightorange}\textbf{267.9$^{\pm 85.6}$} & 1262$^{\pm 96.2}$ \\
\bottomrule
\end{tabular}}
\end{table}

%% file: tables/sensitivity_analysis.tex
\begin{table}[ht]
\centering
\colorlet{lightorange}{orange!25}

\caption{Ablation study on contact reward thresholds. We varied the tolerance ($\epsilon_{\text{tol}}$) and force threshold ($F_{\text{thr}}$) parameters (Eq. 7 and 8).}
\label{tab:ablation_thresholds}
\resizebox{\columnwidth}{!}{
\begin{tabular}{ccccccccc}
\toprule
$\epsilon_{\text{tol}}$ & $F_{\text{thr}}$ & \textbf{SR} $\uparrow$ & \textbf{$E_\mathrm{mpbpe}$} $\downarrow$ & \textbf{$E_\mathrm{g\text{-}mpbpe}$}$\downarrow$& \textbf{$E_\mathrm{mpboe}$} $\downarrow$ & \textbf{$E_\mathrm{mpjpe}$} $\downarrow$ & \textbf{$E_\mathrm{mpope}$} $\downarrow$ & \textbf{$E_\mathrm{mpooe}$} $\downarrow$ \\
\midrule
0.1 & 0 & \cellcolor{lightorange}\textbf{100\%} & 77.6$^{\pm 15.1}$ & 511.7$^{\pm 525.8}$ & 372.7$^{\pm 59.3}$ & {157.3$^{\pm 16.5}$} & 795.7$^{\pm 722.9}$ & 1254$^{\pm 59.4}$ \\
0.05 & 5 & \cellcolor{lightorange}\textbf{100\%} & 68.4$^{\pm 2.4}$ & 388.4$^{\pm 37.6}$ & 334.7$^{\pm 16.9}$ & \cellcolor{lightorange}\textbf{137$^{\pm 1.4}$} & {401.5$^{\pm 37.9}$}& \cellcolor{lightorange}\textbf{1147$^{\pm 227.4}$} \\
0 & 10 & \cellcolor{lightorange}\textbf{100\%} & \cellcolor{lightorange}\textbf{61.3$^{\pm 2.1}$} & \cellcolor{lightorange}\textbf{221.0$^{\pm 74.3}$} & \cellcolor{lightorange}\textbf{319.9$^{\pm 13.0}$} & {145.9$^{\pm 5.5}$} & \cellcolor{lightorange}\textbf{267.9$^{\pm 85.6}$} & 1262$^{\pm 96.2}$ \\
\bottomrule
\end{tabular}}
\vspace{-10pt}
\end{table}

%% file: sections/5_conclusion.tex
\section{Conclusion}

This paper introduced \ours, a novel framework that enables humanoid robots to perform robust whole-body interactions with underactuated objects under visual occlusion by integrating a dynamic-aware world model. By fusing static geometric priors with dynamic proprioceptive feedback, our approach effectively tracks unobserved object states and anticipates environmental changes through higher-order dynamics prediction. This capability was validated in various scenarios, including climbing stairs, dynamic skateboarding, and cart manipulation, demonstrating proactive gait adjustment and balance recovery. Real-world experiments on the Unitree G1 humanoid confirm that our distilled policy achieves high performance, successfully bridging the sim-to-real gap for deployment in sensor-limited environments.

\section{Acknowledgment}

We would like to thank Xiaomi Robotics for providing the experimental site and equipment, Qingfang Li for processing the teaser material, and Shimin Li and Dongfang Li for assisting with the optical motion capture.

%% file: sections/appendix.tex
\appendix\label{app}
\subsection{Dynamics-Aware World Model}
We select the term ``world model'' for our dynamics-aware world model (DWM), as our approach aligns with its fundamental definition in reinforcement learning and control: learning the environment's transition dynamics. DWM explicitly learns the forward transition function by predicting the full state of the object from the previous state $s_{t-1}$ to the current state $s_t$, conditioned on the robot's action and state history. By modeling how the environment responds to robot actions, it serves as a true world model for the interaction system.

\subsection{Dataset Description}
\label{app:dataset}
To achieve robust humanoid-object interaction learning, we constructed a high-fidelity dataset covering diverse interaction scenarios. The data pipeline consists of three main stages: optical motion capture, kinematic retargeting, and physics-based state generation.

\subsubsection{Assets Description}
In our experiments, we employ two primary categories of objects for interaction: fully actuated and underactuated objects.
\begin{itemize}
    \item {Box:} A standard cube with a mass of 1.3 kg. Its external dimensions are 0.36 m (length) × 0.28 m (width) × 0.35 m (height)
    \item {Cart:} The cart has a mass of 13 kg. Its base platform measures 60 cm (width) × 90 cm (length) with a ground clearance of 17 cm, and its handle is 85 cm high.
    \item {Skateboard:} The skateboard has a mass of 3.269 kg, equipped with four wheels (diameter 6.85 cm, thickness 5 cm). Its deck is 24 cm wide with upturned edges, and its ground clearance varies from 12 to 14 cm under load.
\end{itemize}
The composed terrains consist of three fundamental parts: 
\begin{itemize}
    \item {Stair:} The stair has a step height of 10 cm, a tread depth of 30 cm, and a total run length of 90 cm.
    \item {Slope:} The slope has a vertical rise of 30 cm, a slope length of 110 cm, and an inclination angle of approximately 15°.
    \item {Platform:} The platform is a flat, square surface measuring 80 cm × 80 cm. This configuration is designed to replicate common geometric encountered in daily locomotion tasks.
\end{itemize}

\subsubsection{Data Acquisition}

We used an optical motion capture system to record the human demonstration.
\begin{itemize}
    \item {Volunteers:} Participants wore a specialized motion capture suit (equipped with optical markers on key body landmarks) to record whole-body kinematics.
    \item {Interactive Objects:} Markers were affixed to the surface of interactive objects to track their 6-DoF pose (position and orientation) synchronously with the human operator.
    \item {Scenarios:} This dataset covers complex behaviors such as skateboarding, pulling carts, and pushing carts, as well as performing complex operations on multiple objects.
\end{itemize}

\subsubsection{Kinematic Retargeting}
The raw marker data was processed to drive the target humanoid robot (G1) using the PoseLib framework.
\begin{itemize}
    \item {Humanoid Motion:} The human skeleton motion was retargeted to the G1 robot's 29-DoF kinematic structure. This process generated reference trajectories for the robot's root position, root orientation, and 29 joint positions $\bm{q}$.
    \item {Motion of Objects:} The tracked object marker data were cleaned via interpolation of missing frames and transformed into canonical coordinate frames relative to the recording origin, yielding smooth 6-DoF trajectories for the object's root link.
    \item {Alignment:} Coordinate system transformations and spatial offsets were applied to align the human-object interaction workspace with the robot's physical dimensions and the simulation environment.
\end{itemize}

\subsubsection{Physics-Based Motion Representation}
To address the discrepancy between valid interaction and purely kinematic data (lacking physical contact information), we employed a physical simulation method based on the Isaac Gym.

\begin{itemize}
    \item {Digital Twin Simulation:} We reconstructed the scene in simulation by importing the G1 robot and dynamic assets. The robot is driven kinematically by the retargeted reference motion.
    
    \item {Dynamic Objects:} The objects are driven kinematically for trajectory reference.

    \item {Full-State Extraction:}
    \begin{itemize}
        \item \textit{Merged Body State:} We compute a unified state representation that concatenates the robot's $n$ rigid bodies with the interactive $m$ objects into a single ($n+m$)-body array. This provides global positions, linear velocities, quaternions, and angular velocities for the entire system at $50$ Hz.
        \item \textit{Expanded Joint State:} The kinematic joint state is expanded to include object articulations, resulting in a ($29+k$)-DoF vector ($29$ robot joints + $k$ object joints).
    \end{itemize}

    \item {Contact Annotation:}
    \begin{itemize}
        \item \textit{Body Contact:} We heuristically compute contact binary masks by checking geometric proximity between robot links and object meshes.
        \item \textit{Foot Contact:} Ground contact labels are generated based on foot height ($h < 0.14$ m) and velocity thresholds ($v < 1.0$ m/s), incorporating temporal hysteresis for stability.
    \end{itemize}

    \item {Recovery Transitions:} To aid reinforcement learning reset mechanisms, we procedurally generated ``recovery'' transition frames that interpolate between the interaction termination state and a standard ``T-pose'' or standing stance, ensuring valid cycle consistency during training.
\end{itemize}

\input{tables/obs_space}

\subsection{Algorithm Design}

\subsubsection{Observation Space Design}
We adopt an asymmetric observation design, where the student operates under partial observability constraints to ensure deployability, while the teacher leverages privileged simulation information to facilitate training.

\noindent \textbf{Actor Observation (Student)}
The student policy $\pi$ relies solely on onboard sensors and reference information to ensure real-world deployability. The observation vector $o_t \in \mathcal{O}$ primarily consists of:
\begin{itemize}
    \item {Proprioceptive State ($s^{prop}_t$):} This includes a 5-step history of joint positions, joint velocities, root angular velocity, and the projected gravity vector. 
    \item {Reference Motion ($s^{ref}_t$):} To guide the tracking task, the actor receives the reference motion phase $\phi_t$ and a sequence of future reference joint positions.
    \item {Object Point Cloud ($\mathcal{P}$):} To perceive the interaction target, a canonical point cloud of the objects is provided.
\end{itemize}

\noindent \textbf{Critic Observation (Teacher)}
During training, the teacher has access to the privileged state $s^{priv}_t$ to facilitate the learning of complex interaction behaviors. The details are given in Table~\ref{tab:obs_space}.

\subsubsection{Reward Design}
\label{app:reward_details}

The training objective is defined as a weighted sum of individual reward terms, maximizing the cumulative return $R = \sum_{t} \sum_{i} w_i r_{i,t}$. We categorize the rewards into four groups, detailed in Table~\ref{tab:rewards}:
\begin{itemize}
    \item{Motion Tracking:} Encourages the policy to mimic the reference human motion kinematics, including joint angles, body velocities, and root pose.
    \item{Object Interaction:} Critical for task success, this group ensures the object tracks its target trajectory and enforces precise multiple objects contact.
    \item{Foot Constraints:} specific terms designed to ensure physical realism, such as penalizing foot sliding and encouraging stable swing/stance phases.
    \item{Regularization:} Penalizes high-frequency control noise, joint limit violations, and excessive energy consumption to produce smooth and feasible motions.
\end{itemize}

\input{tables/reward_function}
\input{tables/domain_rand}
\input{tables/ppo_hyperparam}
\input{tables/pd_controller}

\subsubsection{Domain Randomization}
To ensure the transferability of our policy to the physical world, we employ extensive domain randomization during training. This technique mitigates the discrepancies between the simulation and reality by modeling physical uncertainties as probabilistic distributions. We randomize a comprehensive set of parameters, ranging from the robot's inertial and actuation properties to the physical characteristics of the interactive objects. Additionally, external perturbations are introduced to enhance stability against unforeseen disturbances. The specific randomization terms and their corresponding sampling ranges are summarized in Table \ref{tab:domain_rand}.

\subsubsection{PPO Hyperparameters and Network Architecture}
We optimize the policy using PPO with the Adam optimizer, utilizing generalized advantage estimation (GAE) and a clipped surrogate objective to ensure stable training. Entropy regularization is applied to encourage exploration. Regarding the architecture, we employ an asymmetric Actor-Critic design implemented as multi-layer perceptrons (MLPs) with ELU activations. The total objective functions as a weighted sum of policy, value, and auxiliary losses. Detailed configurations and network dimensions are listed in Table \ref{tab:ppo_params}.

\subsubsection{Action Space and Low-level Control}

The policy outputs actions $a_t \in \mathbb{R}^{23}$ representing target joint positions $q_{target} = q_t + \Delta q$. Note that this dimension covers the whole-body joints {excluding the 6 wrist degrees of freedom (3 per hand)}, as they are not involved in the interaction tasks. These targets are processed by a low-level proportional-derivative (PD) controller to compute joint torques $\tau$:
\begin{equation}
    \tau = k_p (q_{target} - q_t) - k_d \dot{q}_t
\end{equation}
where $k_p$ and $k_d$ are the stiffness and damping gains. This residual-style control maintains compliant interaction.

The specific PD gains used for the G1 humanoid robot are detailed in Table~\ref{tab:pd_gains}.

\subsection{Experimental Details}

\subsubsection{Experiment Setup}
\label{sec:steup}

\begin{itemize}
    \item {Compute platform}: Each experiment is conducted on a machine with a 24-core Intel i9-13900 CPU running at 5.4GHz, 64 GB of RAM, and a single NVIDIA GeForce RTX 4090 GPU, with Ubuntu 22.04. Each of our models is trained for 8 hours.
    \item {Real robot setup}: We deploy our policies on a Unitree G1 robot. The system consists of an onboard motion control board and an external PC, connected via Ethernet. The control board collects sensor data (IMU and joint states) and transmits it to the PC using the DDS protocol. The PC, running a control loop, estimates the robot's state, maintains observation history, and executes a Finite State Machine (FSM). Inside the FSM, a dedicated RL process performs policy inference in parallel and sends target joint angles (with PD gains) back to the control board, which then issues torque commands to the motors at 500 Hz.
\end{itemize}

\input{tables/sim_task_2}
\input{tables/sim_task_3}

\subsubsection{Evaluation Metrics}\label{app:metric}

\begin{itemize}
    \item{Global Mean Per Body Position Error} ($E_{\rm{g\text{-}mpbpe}}$, mm): The average position error of body parts in global coordinates.
    \begin{equation}
        E_{\rm{g\text{-}mpbpe}} = \mathbb{E} \left[ \big\| \bm{p}_t - \bm{p}_t^{\rm{ref}} \big\|_2 \right].
    \end{equation}

    \item{Root-Relative Mean Per Body Position Error} ($E_{\rm{mpbpe}}$, mm): The average position error of body parts relative to the root position.
    \begin{equation}
        E_{\rm{mpbpe}} = \mathbb{E} \left[ \big\| (\bm{p}_t - \bm{p}_{\rm{root},t}) - (\bm{p}_t^{\rm{ref}} - \bm{p}_{\rm{root},t}^{\rm{ref}}) \big\|_2 \right].
    \end{equation}

    \item{Mean Per Body Orientation Error} ($E_{\rm{mpboe}}$, $10^{-3}$ rad): The average angular error of body part orientations using quaternion distance.
    \begin{equation}
        E_{\rm{mpboe}} = \mathbb{E} \left[ 2 \arccos \left( | \langle \bm{q}_t, \bm{q}_t^{\rm{ref}} \rangle |  \right) \right].
    \end{equation}

    \item{Mean Per Joint Position Error} ($E_{\rm{mpjpe}}$, $10^{-3}$ rad): The average angular error of joint rotations (DOFs).
    \begin{equation}
        E_{\rm{mpjpe}} = \mathbb{E} \left[ \big\| \bm{\theta}_t - \bm{\theta}_t^{\rm{ref}} \big\|_1 \right].
    \end{equation}

    \item{Mean Per Joint Velocity Error} ($E_{\rm{mpjve}}$, $10^{-3}$ rad/frame): The average error of joint angular velocities.
    \begin{equation}
        E_{\rm{mpjve}} = \mathbb{E} \left[ \big\| \Delta \bm{\theta}_t - \Delta \bm{\theta}_t^{\rm{ref}} \big\|_1 \right],
    \end{equation}
    where $\Delta \bm{\theta}_t = \bm{\theta}_t - \bm{\theta}_{t-1}$.

    \item{Mean Per Body Velocity Error} ($E_{\rm{mpbve}}$, mm/frame): The average error in the linear velocities of body parts.
    \begin{equation}
        E_{\rm{mpbve}} = \mathbb{E} \left[ \big\| \Delta \bm{p}_t - \Delta \bm{p}_t^{\rm{ref}} \big\|_2 \right],
    \end{equation}
    where $\Delta \bm{p}_t = \bm{p}_t - \bm{p}_{t-1}$.

    \item{Mean Per Body Acceleration Error} ($E_{\rm{mpbae}}$, mm/frame$^2$): The average error in the linear accelerations of body parts.
    \begin{equation}
        E_{\rm{mpbae}} = \mathbb{E} \left[ \big\| \Delta^2 \bm{p}_t - \Delta^2 \bm{p}_t^{\rm{ref}} \big\|_2 \right],
    \end{equation}
    where $\Delta^2 \bm{p}_t = \Delta \bm{p}_t - \Delta \bm{p}_{t-1}$.
    
    \item{Mean Per Object Position Error} ($E_{\rm{mpope}}$, mm): The average position error in the global coordinates of object parts.
    \begin{equation}
        E_{\rm{mpope}} = \mathbb{E} \left[ \big\| \bm{p}_t^{\rm{obj}} - \bm{p}_t^{\rm{obj, ref}} \big\|_2 \right].
    \end{equation}

    \item{Mean Per Object Orientation Error} ($E_{\rm{mpooe}}$, $10^{-3}$ rad): The average angular error in object orientations.
    \begin{equation}
        E_{\rm{mpooe}} = \mathbb{E} \left[ 2 \arccos \left( | \langle \bm{q}_t^{\rm{obj}}, \bm{q}_t^{\rm{obj, ref}} \rangle | \right) \right].
    \end{equation}

    \item{Mean Per Object Velocity Error} ($E_{\rm{mpove}}$, mm/frame): The average error in object linear velocities.
    \begin{equation}
        E_{\rm{mpove}} = \mathbb{E} \left[ \big\| \Delta \bm{p}_t^{\rm{obj}} - \Delta \bm{p}_t^{\rm{obj, ref}} \big\|_2 \right],
    \end{equation}
    where $\Delta \bm{p}_t^{\rm{obj}} = \bm{p}_t^{\rm{obj}} - \bm{p}_{t-1}^{\rm{obj}}$.

    \item{Mean Per Object Acceleration Error} ($E_{\rm{mpoae}}$, mm/frame$^2$): The average error of object linear accelerations.
    \begin{equation}
        E_{\rm{mpoae}} = \mathbb{E} \left[ \big\| \Delta^2 \bm{p}_t^{\rm{obj}} - \Delta^2 \bm{p}_t^{\rm{obj, ref}} \big\|_2 \right],
    \end{equation}
    where $\Delta^2 \bm{p}_t^{\rm{obj}} = \Delta \bm{p}_t^{\rm{obj}} - \Delta \bm{p}_{t-1}^{\rm{obj}}$.
\end{itemize}

\subsection{Additional Experimental Results}

\subsubsection{Inference Latency}
Real-time feasibility is paramount for agile interaction. 
By utilizing a sparse canonical template instead of dense raw scans, the {explicit geometric projection} (Eq.~\ref{eq:geo_proj}) is simplified to efficient vectorized matrix operations.
The benchmarks on Unitree G1 confirm that the total inference latency averages {0.17 ms}, consuming less than 1\% of the control period of 20 ms (50 Hz). 
This negligible overhead guarantees an instantaneous response capability for high-speed tasks such as skateboarding.

\subsubsection{Ablation Study for Sequential Interaction}

Table \ref{tab:sim_case2_full} evaluates the performance in sequential tasks.

\noindent \textbf{Failure of Baselines in Sequential Tasks:} The complexity of these tasks causes a severe accumulation of errors in baseline methods. \ma and \mb fail completely on the ``Pull Cart w/ Box'' task, dropping to a {0\% success rate}. This indicates that standard policies without explicit future prediction struggle to coordinate the transition between manipulating the box and controlling the cart.

\noindent \textbf{Mitigating Sequential Drift:} In contrast, \ours maintains a {high success rate} across both ``Pull'' and ``Push'' tasks. Our dynamics-aware world model effectively mitigates the drift that typically accumulates over long horizons. This is evidenced by the significantly lower global body position error in the ``Pull'' task.

\noindent \textbf{Precision in Complex Interaction:} In the ``Push Cart w/ Box'' task, \ours achieves the best performance across almost all metrics. Notably, it records the lowest object orientation error, demonstrating that our method can stably steer the cart-box system, whereas other methods struggle to maintain the heading of the coupled objects.

\subsubsection{Ablation Study for Multi-terrain Interaction}

Table \ref{tab:sim_case3_full} compares the performance across different terrains.

\noindent \textbf{Robustness Comparison:} On simple terrains, all methods achieve a high success rate. However, on the most challenging slope-stair terrain, \ma and \mb fail to adapt to rapid height changes. In contrast, {\ours maintains a high success rate} across all scenarios, showing superior robustness against environmental complexity.

\noindent \textbf{Tracking Accuracy:} \ours consistently achieves lower tracking errors than the baselines. On the slope-stair terrain, our method outperforms \md significantly in global body position error and object orientation error. This indicates that our policy generates more stable whole-body motions and precise object manipulation.

%% file: tables/obs_space.tex
\begin{table*}[h]
\centering
\caption{Details of the Observation Space and Privileged Information.}
\label{tab:obs_space}
\renewcommand{\arraystretch}{1.3}
\begin{tabular}{l|l|c|l}
\toprule
\textbf{Category} & \textbf{Observation Term} & \textbf{Noise ($\sigma$)} & \textbf{Description} \\ 
\midrule
\multirow{8}{*}{\textbf{Proprioception}} 
& Base Angular Velocity & 0.05 & History of base angular velocity in base frame (steps: [0]). \\
& Projected Gravity & 0.05 & History of gravity vector projected to base frame (steps: [0]). \\
& Joint Positions & 0.015 & History of joint positions (steps: [0, 1, 2, 3, 4, 8]). \\
& Previous Actions & - & Joint target history from previous 3 steps. \\
& Ref. Motion Phase & - & Phase variable $\phi$ of the reference motion. \\
& Ref. Joint Positions & - & Future reference joint positions. \\
& Ref. Body Pos (Local) & - & Future reference body positions in robot local frame. \\
& Ref. Contact 1 Pos & 0.01 & Reference contact positions on Object 1. \\
& Ref. Contact 2 Pos & 0.01 & Reference contact positions on Object 2 (if enabled). \\
& Objects Point Cloud & - & Canonical point clouds of Object 1 \& Object 2 (if enabled). \\
\midrule
\multirow{17}{*}{\textbf{Privileged}} 
& \textit{Clean Proprioception} & 0.0 & Noise-free history (steps: 0-8) of all proprioceptive states. \\
& Root Linear Velocity & 0.0 & Linear velocity of the robot base in local frame. \\
& Body Velocity & 0.0 & Linear velocity of key bodies (e.g., ankles). \\
& Body Height & 0.0 & Height of pelvis, torso, and feet relative to ground. \\
& Ref. Root State (Global) & 0.0 & Future reference root position/orientation in global frame. \\
& Ref. Diff (Local) & 0.0 & Difference between current and ref (pos, ori, ...) in local frame. \\
& Obj. 1 State & 0.01 & Relative state (pos, ori, ...) of Object 1 in local frame. \\
& Obj. 2 State & 0.01 & Relative state (pos, ori, ...) of Object 2 (if enabled) in local frame. \\
& Applied Forces & - & Real applied actions and torques. \\
& \multicolumn{3}{l}{\textit{Dynamics Randomization Parameters (Implicitly observed by Critic):}} \\
& \multicolumn{3}{l}{\quad Body/Object Mass, Friction, Restitution, Object Scale, Joint Armature/Damping.} \\
\bottomrule
\end{tabular}
\end{table*}

%% file: tables/reward_function.tex
\begin{table*}[ht]
\centering
\caption{Reward Functions for the Humanoid-Object Interaction Task.}
\label{tab:rewards}
\renewcommand{\arraystretch}{1.3}
\begin{tabular}{llcl}
\toprule
\textbf{Term} & \textbf{Expression} & \textbf{Weight} & \textbf{Description} \\ 
\midrule
\multicolumn{4}{l}{\textit{{(a) Motion Tracking Reward}}} \\
\hdashline
Joint Position & $\exp(-\| \bm{q} - \bm{q}^{\text{ref}} \|^2 / \sigma_q)$ & 0.5 & Tracks reference joint positions. \\
Joint Velocity & $\exp(-\| \dot{\bm{q}} - \dot{\bm{q}}^{\text{ref}} \|^2 / \sigma_{\dot{q}})$ & 0.5 & Tracks reference joint velocities. \\
Upper Body Pos. & $\exp(-\| \bm{p}_{\text{up}} - \bm{p}_{\text{up}}^{\text{ref}} \|^2 / \sigma_p)$ & 0.5 & Tracks pos. of shoulder, elbow, wrist. \\
Upper Body Ori. & $\exp(-\| \bm{\theta}_{\text{up}} \ominus \bm{\theta}_{\text{up}}^{\text{ref}} \|^2 / \sigma_\theta)$ & 0.5 & Tracks ori. of upper body links. \\
Lower Body Pos. & $\exp(-\| \bm{p}_{\text{low}} - \bm{p}_{\text{low}}^{\text{ref}} \|^2 / \sigma_p)$ & 0.5 & Tracks pos. of hip, knee, ankle links. \\
Lower Body Ori. & $\exp(-\| \bm{\theta}_{\text{low}} \ominus \bm{\theta}_{\text{low}}^{\text{ref}} \|^2 / \sigma_\theta)$ & 0.5 & Tracks ori. of lower body links. \\
Root Position & $\exp(-\| \bm{p}_{\text{root}} - \bm{p}_{\text{root}}^{\text{ref}} \|^2 / \sigma_p)$ & 0.5 & Tracks pelvis position in world frame. \\
Root Orientation & $\exp(-\| \bm{\theta}_{\text{root}} \ominus \bm{\theta}_{\text{root}}^{\text{ref}} \|^2 / \sigma_\theta)$ & 0.5 & Tracks pelvis orientation. \\
Body Lin/Ang Vel. & $\exp(-\| \bm{v} - \bm{v}^{\text{ref}} \|^2 / \sigma_v)$ & 0.5 & Tracks lin/ang velocities of bodies. \\

\midrule
\multicolumn{4}{l}{\textit{{(b) Object Interaction Reward}}} \\
\hdashline
Object Position & $\exp(-\| \bm{p}_{\text{obj}} - \bm{p}_{\text{obj}}^{\text{ref}} \|^2 / \sigma_p)$ & 1.0 & Tracks global position of object(s). \\
Object Orientation & $\exp(-\| \bm{\theta}_{\text{obj}} \ominus \bm{\theta}_{\text{obj}}^{\text{ref}} \|^2 / \sigma_\theta)$ & 1.0 & Tracks global orientation of object(s). \\
Multiple Objects Contact & $ \frac{1}{|\mathcal{O}|} \sum_{o \in \mathcal{O}} \left( \frac{1}{|\mathcal{E}_o|} \sum_{e \in \mathcal{E}_o} \mathbb{I}_{o,e} \cdot r_{\text{pos}}^{o,e} \cdot r_{\text{force}}^{o,e} \right)$ & 1.0 & Align active body-object pairs \& forces. \\

\midrule
\multicolumn{4}{l}{\textit{{(c) Foot Constraints}}} \\
\hdashline
Feet Air Time & $\exp( \text{clip}(t_{\text{air}} - t_{\text{thr}}) / \sigma ) \cdot \mathbb{I}_{\text{step}}$ & 0.5 & Encourages longer swing phases. \\
Feet Slip & $-\| \bm{v}_{\text{foot}}^{xy} \| \cdot \mathbb{I}_{\text{ground}}$ & 0.5 & Penalizes sliding velocity when grounded. \\
Feet Contact Match & $\exp(-\| \mathbb{I}_{\text{con}}^{\text{real}} - \mathbb{I}_{\text{con}}^{\text{ref}} \|^2 / \sigma)$ & 0.5 & Matches reference contact states. \\
Feet Air Lift & $-\sum (\bm{h}_{\text{foot}} < h_{\text{min}}) \cdot \mathbb{I}_{\text{swing}}$ & 0.5 & Penalizes tripping during swing. \\
Impact Force & $-\| \bm{F}_{\text{impact}} \|^2$ & 1.0 & Penalizes large impact forces. \\

\midrule
\multicolumn{4}{l}{\textit{{(d) Regularization}}} \\
\hdashline
Action Rate & $-\| \bm{a}_t - \bm{a}_{t-1} \|^2$ & 0.1 & Penalizes rapid action changes. \\
Joint Velocity L2 & $-\| \dot{\bm{q}} \|^2$ & 5e-4 & Penalizes high velocity (energy). \\
Joint Limits & $-\sum \text{clip}(\bm{q} - \bm{q}_{\text{limit}})$ & 10.0 & Penalizes exceeding physical limits. \\
Torque Limits & $-\sum \text{clip}(\bm{\tau} - \bm{\tau}_{\text{limit}})$ & 0.01 & Penalizes torque saturation. \\
Survival & $1.0$ & 1.0 & Reward for not terminating early. \\
\bottomrule
\end{tabular}
\end{table*}

%% file: tables/domain_rand.tex
\begin{table}[h]
\centering
\caption{Domain Randomization Parameters for Robot and Objects.}
\label{tab:domain_rand}
\renewcommand{\arraystretch}{1.3}
\begin{tabular}{llc}
\toprule
\textbf{Category} & \textbf{Parameter} & \textbf{Range / Distribution} \\ 
\midrule
\multicolumn{3}{l}{\textit{\textbf{Robot Dynamics}}} \\
\hdashline
\multirow{4}{*}{Properties} 
& Link Mass Scale & $\mathcal{U}(0.9, 1.1) \times \text{default}$ \\
& Center of Mass Offset & $\mathcal{U}(-0.02, 0.02)$ m \\
& Static Friction  & $\mathcal{U}(0.3, 1.6)$ \\
& Dynamic Friction  & $\mathcal{U}(0.3, 1.2)$ \\
\midrule
\multirow{4}{*}{Actuation} 
& Joint Position Offset & $\mathcal{U}(-0.01, 0.01)$ rad \\
& Motor Stiffness Scale & $\mathcal{U}(0.9, 1.1)$ \\
& Motor Damping Scale & $\mathcal{U}(0.9, 1.1)$ \\
& Action Delay & $\mathcal{U}[40, 120]$ ms \\

\midrule
\multicolumn{3}{l}{\textit{\textbf{Object Interaction}}} \\
\hdashline
\multirow{3}{*}{{Surface}} 
& Dynamic Friction & $\mathcal{U}(0.3, 0.8)$ \\
& Static-to-Dynamic Ratio & $\mathcal{U}(1.0, 2.0)$ \\
& Restitution & $\mathcal{U}(0.0, 0.2)$ \\
\midrule
\multirow{2}{*}{{Box}} 
& Mass & $\mathcal{U}(1.0, 2.0)$ kg \\
& Scale & $\mathcal{U}(0.9, 1.1)$ \\
\midrule
\multirow{5}{*}{{Cart}} 
& Body Mass & $\mathcal{U}(11.0, 13.0)$ kg \\
& Wheel Mass & $\mathcal{U}(0.2, 0.4)$ kg \\
& Wheel Joint Friction & $\mathcal{U}(0.01, 0.1)$ N$\cdot$m \\
& Wheel Joint Damping & $\mathcal{U}(0.01, 0.1)$ N$\cdot$m$\cdot$s/rad \\
& Scale & $\mathcal{U}(0.9, 1.1)$ \\
\midrule
\multirow{5}{*}{{Skateboard}} 
& Body Mass & $\mathcal{U}(2.0, 5.0)$ kg \\
& Wheel Mass & $\mathcal{U}(0.1, 0.2)$ kg \\
& Wheel Armature & $\mathcal{U}(0.0, 1\text{e-}4)$ kg$\cdot$m$^2$ \\
& Wheel Joint Damping & $\mathcal{U}(0.0, 1\text{e-}3)$ N$\cdot$m$\cdot$s/rad \\
& Scale  & $\mathcal{U}(0.9, 1.1)$ \\
\midrule
{Slope / Stair}
& Scale  & $\mathcal{U}(0.98, 1.02)$ \\

\midrule
\multicolumn{3}{l}{\textit{\textbf{External Perturbation}}} \\
\hdashline
\multirow{2}{*}{{Push}} 
& Push Force& $\mathcal{U}(0.2, 0.5) \times \text{weight}$ \\
& Push Min Interval & 2s \\
\bottomrule
\end{tabular}
\end{table}

%% file: tables/ppo_hyperparam.tex
\begin{table}[h]
\centering
\caption{Hyperparameters related to PPO and Network Architecture.}
\label{tab:ppo_params}
\renewcommand{\arraystretch}{1.3}
\begin{tabular}{lc}
\toprule
\textbf{Hyperparameter} & \textbf{Value} \\ 
\midrule
Optimizer & Adam \\
Number of Environments & 4096 \\
Rollout Steps (Horizon) & 32 \\
Mini-batches & 8 \\
Learning Epochs & 3 \\
Discount Factor ($\gamma$) & 0.99 \\
GAE Parameter ($\lambda$) & 0.95 \\
Clip Parameter ($\epsilon$) & 0.2 \\
Entropy Coefficient & 0.001 \\
Max Gradient Norm & 1.0 \\
Desired KL & 0.01 \\
Learning Rate & $3 \times 10^{-4}$ \\
Initial Noise Std & 1.0 \\
\midrule
\multicolumn{2}{l}{\textit{Loss Coefficients}} \\
\hdashline
Value Loss Coefficient ($\lambda_{\text{value}}$) & 1.0 \\
Object Loss Coefficient ($\lambda_{\text{obj}}$) & 1.0 \\
Privileged Loss Coefficient ($\lambda_{\text{priv}}$) & 1.0 \\
Distillation Loss Coefficient ($\lambda_{\text{distill}}$) & 1.0 \\
\midrule
\multicolumn{2}{l}{\textit{Network Architecture}} \\
\hdashline
Actor MLP Size & $[512, 256, 256]$ \\
Critic MLP Size & $[512, 256, 128]$ \\
Adapter MLP Size & $[256, 256]$ \\
Activation Function & ELU \\
\bottomrule
\end{tabular}
\end{table}

%% file: tables/pd_controller.tex
\begin{table}[ht]
    \centering
    \caption{PD controller gains for the G1 Humanoid.}
    \label{tab:pd_gains}
    \renewcommand{\arraystretch}{1.3}
    \begin{tabular}{lcc}
        \toprule
        \textbf{Joint Name} & \textbf{Stiffness ($k_p$)} & \textbf{Damping ($k_d$)} \\
        \midrule
        \multicolumn{3}{l}{\textit{Legs}} \\
        \hdashline
        Left/Right Hip Pitch/Yaw    & 40.18 & 2.558 \\
        Left/Right Hip Roll         & 99.10 & 6.309 \\
        Left/Right Knee             & 99.10 & 6.309 \\
        Left/Right Ankle Pitch/Roll & 28.50 & 1.814 \\
        \midrule
        \multicolumn{3}{l}{\textit{Waist}} \\
        \hdashline
        Waist Yaw                   & 40.18 & 2.558 \\
        Waist Roll/Pitch            & 28.50 & 1.814 \\
        \midrule
        \multicolumn{3}{l}{\textit{Arms}} \\
        \hdashline
        Left/Right Shoulder Pitch/Roll/Yaw & 14.25 & 0.9072 \\
        Left/Right Elbow Pitch/Roll        & 14.25 & 0.9072 \\
        Left/Right Wrist Roll/Pitch        & 16.78 & 1.068  \\
        \bottomrule
    \end{tabular}
\end{table}

%% file: tables/sim_task_2.tex
\begin{table*}[h]
\centering
\colorlet{lightorange}{orange!25}
\colorlet{lighterorange}{orange!12}

\caption{Ablation results for \textbf{Sequential Interaction}. The complexity of sequentially loading and manipulating objects causes error accumulation in \ma. \ma and \mb fail completely on ``Pull Cart w/ Box''. \textbf{\ours} maintains a high success rate across both tasks and achieves the lowest object orientation error in the ``Push'' phase, confirming that the dynamic-aware world model effectively mitigates drift over long horizons.}
\label{tab:sim_case2_full}
\resizebox{\textwidth}{!}{
\begin{tabular}{llclllllllllll}
\toprule
& & & \multicolumn{7}{c}{\textbf{Robot State Metrics}} & \multicolumn{4}{c}{\textbf{Object State Metrics}} \\
\cmidrule(lr){4-10} \cmidrule(lr){11-14}
\textbf{Task} & \textbf{Method} & \textbf{SR} $\uparrow$ & \textbf{$E_\mathrm{mpbpe}$}$\downarrow$ & \textbf{$E_\mathrm{g\text{-}mpbpe}$}$\downarrow$ & \textbf{$E_\mathrm{mpboe}$}$\downarrow$ & \textbf{$E_\mathrm{mpjpe}$}$\downarrow$ & \textbf{$E_\mathrm{mpbve}$}$\downarrow$ & \textbf{$E_\mathrm{mpbae}$}$\downarrow$ & \textbf{$E_\mathrm{mpjve}$}$\downarrow$ & \textbf{$E_\mathrm{mpope}$}$\downarrow$ & \textbf{$E_\mathrm{mpooe}$}$\downarrow$ & \textbf{$E_\mathrm{mpove}$}$\downarrow$ & \textbf{$E_\mathrm{mpoae}$}$\downarrow$ \\
\midrule

\multirow{5}{*}{\makecell[l]{Pull Cart\\w/ Box}} 
& \ma & 0\% & 117.9$^{\pm 31.3}$ & 354.7$^{\pm 93.0}$ & 590.8$^{\pm 124}$ & 162.7$^{\pm 14.1}$ & 5.67$^{\pm 0.24}$ & \cellcolor{lighterorange}4.64$^{\pm 0.04}$ & 12.21$^{\pm 0.15}$ & 501.6$^{\pm 47.3}$ & \cellcolor{lighterorange}454.4$^{\pm 13.5}$ & 4.63$^{\pm 1.04}$ & \cellcolor{lighterorange}2.87$^{\pm 0.01}$ \\
& \mb & 0\% & 138.7$^{\pm 2.5}$ & 444.8$^{\pm 118}$ & 780.9$^{\pm 75.7}$ & 155.8$^{\pm 1.4}$ & 5.25$^{\pm 0.09}$ & \cellcolor{lightorange}\textbf{4.55$^{\pm 0.02}$} & \cellcolor{lightorange}\textbf{10.57$^{\pm 0.25}$} & 550.6$^{\pm 218}$ & \cellcolor{lightorange}\textbf{438.7$^{\pm 77.6}$} & 4.83$^{\pm 0.92}$ & \cellcolor{lighterorange}2.87$^{\pm 0.01}$ \\
& \mc & 100\% & \cellcolor{lightorange}\textbf{59.7$^{\pm 2.7}$} & \cellcolor{lighterorange}204.7$^{\pm 21.2}$ & \cellcolor{lightorange}\textbf{333.5$^{\pm 15.0}$} & \cellcolor{lightorange}\textbf{130.0$^{\pm 2.0}$} & \cellcolor{lightorange}\textbf{5.18$^{\pm 0.14}$} & \cellcolor{lighterorange}4.64$^{\pm 0.10}$ & \cellcolor{lighterorange}11.88$^{\pm 0.42}$ & \cellcolor{lightorange}\textbf{300.6$^{\pm 26.1}$} & 499.3$^{\pm 66.3}$ & \cellcolor{lightorange}\textbf{4.06$^{\pm 0.39}$} & 2.88$^{\pm 0.01}$ \\
& \md & 100\% & 69.8$^{\pm 2.0}$ & 302.2$^{\pm 65.9}$ & 376.6$^{\pm 14.5}$ & 138.4$^{\pm 0.9}$ & 5.63$^{\pm 0.13}$ & 4.75$^{\pm 0.02}$ & 12.45$^{\pm 0.19}$ & 414.7$^{\pm 64.7}$ & 635.5$^{\pm 68.1}$ & 4.47$^{\pm 0.37}$ & 2.88$^{\pm 0.01}$ \\
& \textbf{\ours} & 100\% & \cellcolor{lighterorange}64.4$^{\pm 2.1}$ & \cellcolor{lightorange}\textbf{196.5$^{\pm 29.5}$} & \cellcolor{lighterorange}343.7$^{\pm 20.3}$ & \cellcolor{lighterorange}135.8$^{\pm 5.2}$ & \cellcolor{lighterorange}5.24$^{\pm 0.14}$ & 4.65$^{\pm 0.06}$ & 12.17$^{\pm 0.46}$ & \cellcolor{lighterorange}332.9$^{\pm 30.0}$ & 561.3$^{\pm 22.6}$ & \cellcolor{lighterorange}4.35$^{\pm 0.26}$ & \cellcolor{lightorange}\textbf{2.86$^{\pm 0.01}$} \\
\midrule

\multirow{5}{*}{\makecell[l]{Push Cart\\w/ Box}} 
& \ma & 0\% & 99.7$^{\pm 39.2}$ & 520.7$^{\pm 240}$ & 605.0$^{\pm 263}$ & 167.0$^{\pm 9.6}$ & 5.67$^{\pm 0.64}$ & 3.84$^{\pm 0.27}$ & 12.23$^{\pm 1.42}$ & 536.7$^{\pm 171}$ & \cellcolor{lighterorange}362.2$^{\pm 78.5}$ & 4.43$^{\pm 1.01}$ & 3.03$^{\pm 0.03}$ \\
& \mb & 80\% & 66.2$^{\pm 2.0}$ & 451.9$^{\pm 69.3}$ & 393.6$^{\pm 13.4}$ & 175.2$^{\pm 3.5}$ & 4.97$^{\pm 0.15}$ & 3.52$^{\pm 0.11}$ & 11.42$^{\pm 0.52}$ & 499.3$^{\pm 79.4}$ & 422.3$^{\pm 41.9}$ & 4.22$^{\pm 0.28}$ & 3.02$^{\pm 0.01}$ \\
& \mc & 100\% & 68.3$^{\pm 5.0}$ & 490.2$^{\pm 46.3}$ & 407.7$^{\pm 21.9}$ & 156.2$^{\pm 6.1}$ & 5.00$^{\pm 0.24}$ & \cellcolor{lighterorange}3.48$^{\pm 0.09}$ & 11.68$^{\pm 0.51}$ & 519.0$^{\pm 47.1}$ & 386.3$^{\pm 72.6}$ & \cellcolor{lighterorange}4.06$^{\pm 0.22}$ & \cellcolor{lighterorange}3.01$^{\pm 0.01}$ \\
& \md & 100\% & \cellcolor{lightorange}\textbf{58.2$^{\pm 1.0}$} & \cellcolor{lighterorange}360.2$^{\pm 9.1}$ & \cellcolor{lightorange}\textbf{324.8$^{\pm 6.0}$} & \cellcolor{lighterorange}141.6$^{\pm 2.0}$ & \cellcolor{lighterorange}4.95$^{\pm 0.13}$ & 3.54$^{\pm 0.08}$ & \cellcolor{lighterorange}11.39$^{\pm 0.42}$ & \cellcolor{lighterorange}491.3$^{\pm 68.1}$ & 419.1$^{\pm 63.7}$ & 4.11$^{\pm 0.12}$ & 3.02$^{\pm 0.02}$ \\
& \textbf{\ours} & 100\% & \cellcolor{lighterorange}60.2$^{\pm 3.2}$ & \cellcolor{lightorange}\textbf{326.5$^{\pm 11.3}$} & \cellcolor{lighterorange}327.7$^{\pm 4.3}$ & \cellcolor{lightorange}\textbf{140.8$^{\pm 0.8}$} & \cellcolor{lightorange}\textbf{4.75$^{\pm 0.03}$} & \cellcolor{lightorange}\textbf{3.42$^{\pm 0.02}$} & \cellcolor{lightorange}\textbf{10.73$^{\pm 0.05}$} & \cellcolor{lightorange}\textbf{386.7$^{\pm 18.0}$} & \cellcolor{lightorange}\textbf{326.9$^{\pm 7.3}$} & \cellcolor{lightorange}\textbf{4.00$^{\pm 0.13}$} & \cellcolor{lightorange}\textbf{3.00$^{\pm 0.01}$} \\
\bottomrule
\end{tabular}}
\end{table*}

%% file: tables/sim_task_3.tex
\begin{table*}[ht]
\centering
\colorlet{lightorange}{orange!25}
\colorlet{lighterorange}{orange!12}

\caption{Ablation results for \textbf{Multi-terrain Interaction}. \textbf{\ours} achieves the most consistent success rate across all terrains, while \md demonstrates superior stability in velocity and acceleration metrics on complex stair terrains due to the explicit geometric projection.}
\label{tab:sim_case3_full}
\resizebox{\textwidth}{!}{
\begin{tabular}{llclllllllllll}
\toprule
& & & \multicolumn{7}{c}{\textbf{Robot State Metrics}} & \multicolumn{4}{c}{\textbf{Object State Metrics}} \\
\cmidrule(lr){4-10} \cmidrule(lr){11-14}
\textbf{Task} & \textbf{Method} & \textbf{SR} $\uparrow$ & \textbf{$E_\mathrm{mpbpe}$} & \textbf{$E_\mathrm{g\text{-}mpbpe}$}$\downarrow$ & \textbf{$E_\mathrm{mpboe}$}$\downarrow$ & \textbf{$E_\mathrm{mpjpe}$}$\downarrow$ & \textbf{$E_\mathrm{mpbve}$}$\downarrow$ & \textbf{$E_\mathrm{mpbae}$}$\downarrow$ & \textbf{$E_\mathrm{mpjve}$}$\downarrow$ & \textbf{$E_\mathrm{mpope}$}$\downarrow$ & \textbf{$E_\mathrm{mpooe}$}$\downarrow$ & \textbf{$E_\mathrm{mpove}$}$\downarrow$ & \textbf{$E_\mathrm{mpoae}$}$\downarrow$ \\
\midrule

\multirow{5}{*}{\makecell[l]{Carry\\Box}} 
& \ma & 100\% & \cellcolor{lighterorange}61.8$^{\pm 4.8}$ & \cellcolor{lighterorange}166.3$^{\pm 10.3}$ & 323.6$^{\pm 16.7}$ & \cellcolor{lightorange}\textbf{181.6$^{\pm 5.0}$} & 5.34$^{\pm 0.23}$ & 3.34$^{\pm 0.07}$ & 11.08$^{\pm 0.50}$ & 180.1$^{\pm 12.0}$ & \cellcolor{lightorange}\textbf{129.0$^{\pm 7.3}$} & 7.23$^{\pm 0.09}$ & 1.84$^{\pm 0.05}$ \\
& \mb & 100\% & \cellcolor{lightorange}\textbf{59.3$^{\pm 4.4}$} & \cellcolor{lightorange}\textbf{155.7$^{\pm 17.6}$} & \cellcolor{lightorange}\textbf{315.9$^{\pm 19.3}$} & 194.5$^{\pm 1.4}$ & 5.78$^{\pm 0.12}$ & 3.72$^{\pm 0.08}$ & 13.49$^{\pm 0.28}$ & \cellcolor{lightorange}\textbf{161.8$^{\pm 24.6}$} & \cellcolor{lighterorange}149.9$^{\pm 22.6}$ & 6.92$^{\pm 0.22}$ & 2.23$^{\pm 0.11}$ \\
& \mc & 100\% & 73.2$^{\pm 5.3}$ & 220.2$^{\pm 52.1}$ & 340.2$^{\pm 24.1}$ & 198.2$^{\pm 2.9}$ & 5.19$^{\pm 0.06}$ & 3.32$^{\pm 0.06}$ & 11.20$^{\pm 0.28}$ & 197.5$^{\pm 50.4}$ & 517.4$^{\pm 45.8}$ & 7.24$^{\pm 0.35}$ & 1.99$^{\pm 0.08}$ \\
& \md & 100\% & 69.4$^{\pm 1.6}$ & 221.8$^{\pm 13.8}$ & \cellcolor{lighterorange}318.7$^{\pm 6.9}$ & 209.3$^{\pm 1.2}$ & \cellcolor{lightorange}\textbf{4.82$^{\pm 0.12}$} & \cellcolor{lightorange}\textbf{3.18$^{\pm 0.07}$} & \cellcolor{lightorange}\textbf{10.20$^{\pm 0.18}$} & 184.6$^{\pm 22.5}$ & 258.3$^{\pm 8.7}$ & \cellcolor{lightorange}\textbf{6.26$^{\pm 0.25}$} & \cellcolor{lighterorange}1.71$^{\pm 0.04}$ \\
& \textbf{\ours} & 100\% & 64.3$^{\pm 2.9}$ & 173.2$^{\pm 34.4}$ & 320.4$^{\pm 18.7}$ & \cellcolor{lighterorange}182.5$^{\pm 0.8}$ & \cellcolor{lighterorange}5.17$^{\pm 0.05}$ & \cellcolor{lighterorange}3.23$^{\pm 0.04}$ & \cellcolor{lighterorange}10.29$^{\pm 0.15}$ & \cellcolor{lighterorange}176.0$^{\pm 30.9}$ & 169.2$^{\pm 5.5}$ & \cellcolor{lighterorange}6.42$^{\pm 0.23}$ & \cellcolor{lightorange}\textbf{1.62$^{\pm 0.02}$} \\
\midrule

\multirow{5}{*}{\makecell[l]{w/ Stair}} 
& \ma & 100\% & 66.1$^{\pm 4.6}$ & 234.7$^{\pm 28.8}$ & 351.5$^{\pm 15.5}$ & 166.5$^{\pm 10.8}$ & 4.95$^{\pm 0.20}$ & 4.14$^{\pm 0.06}$ & 12.02$^{\pm 0.36}$ & \cellcolor{lighterorange}148.2$^{\pm 20.9}$ & 193.9$^{\pm 100.8}$ & 3.88$^{\pm 0.40}$ & 2.57$^{\pm 0.09}$ \\
& \mb & 100\% & 55.3$^{\pm 1.3}$ & \cellcolor{lighterorange}218.0$^{\pm 11.4}$ & \cellcolor{lighterorange}300.6$^{\pm 9.8}$ & 132.3$^{\pm 1.5}$ & 4.34$^{\pm 0.05}$ & 3.95$^{\pm 0.02}$ & \cellcolor{lighterorange}9.41$^{\pm 0.17}$ & \cellcolor{lightorange}\textbf{138.4$^{\pm 35.5}$} & 390.9$^{\pm 101.2}$ & 3.79$^{\pm 0.23}$ & 2.40$^{\pm 0.06}$ \\
& \mc & 100\% & 57.8$^{\pm 4.4}$ & \cellcolor{lightorange}\textbf{216.4$^{\pm 9.9}$} & 317.3$^{\pm 30.3}$ & \cellcolor{lighterorange}132.1$^{\pm 0.5}$ & 4.39$^{\pm 0.02}$ & 3.97$^{\pm 0.03}$ & 9.61$^{\pm 0.01}$ & 149.5$^{\pm 29.0}$ & 294.8$^{\pm 145.7}$ & 3.73$^{\pm 0.06}$ & 2.38$^{\pm 0.02}$ \\
& \md & 100\% & \cellcolor{lighterorange}53.7$^{\pm 1.0}$ & 283.0$^{\pm 30.8}$ & 302.9$^{\pm 1.7}$ & 132.6$^{\pm 1.3}$ & \cellcolor{lightorange}\textbf{3.91$^{\pm 0.01}$} & \cellcolor{lightorange}\textbf{3.84$^{\pm 0.03}$} & \cellcolor{lightorange}\textbf{8.91$^{\pm 0.11}$} & 168.9$^{\pm 7.9}$ & \cellcolor{lighterorange}82.0$^{\pm 2.6}$ & \cellcolor{lightorange}\textbf{3.17$^{\pm 0.05}$} & \cellcolor{lightorange}\textbf{2.32$^{\pm 0.04}$} \\
& \textbf{\ours} & 100\% & \cellcolor{lightorange}\textbf{53.5$^{\pm 0.9}$} & 281.5$^{\pm 72.9}$ & \cellcolor{lightorange}\textbf{290.5$^{\pm 0.8}$} & \cellcolor{lightorange}\textbf{127.5$^{\pm 2.5}$} & \cellcolor{lighterorange}4.34$^{\pm 0.04}$ & \cellcolor{lighterorange}3.95$^{\pm 0.03}$ & 9.67$^{\pm 0.08}$ & 154.5$^{\pm 29.8}$ & \cellcolor{lightorange}\textbf{71.0$^{\pm 7.4}$} & \cellcolor{lighterorange}3.41$^{\pm 0.03}$ & \cellcolor{lighterorange}2.40$^{\pm 0.02}$ \\
\midrule

\multirow{5}{*}{\makecell[l]{w/ Slope}} 
& \ma & 100\% & \cellcolor{lightorange}\textbf{54.6$^{\pm 1.1}$} & \cellcolor{lighterorange}186.6$^{\pm 14.4}$ & \cellcolor{lighterorange}317.7$^{\pm 3.6}$ & 138.4$^{\pm 2.4}$ & \cellcolor{lightorange}\textbf{3.72$^{\pm 0.04}$} & \cellcolor{lightorange}\textbf{3.30$^{\pm 0.03}$} & \cellcolor{lightorange}\textbf{8.78$^{\pm 0.08}$} & \cellcolor{lighterorange}105.4$^{\pm 5.1}$ & 108.4$^{\pm 17.0}$ & \cellcolor{lighterorange}3.06$^{\pm 0.06}$ & \cellcolor{lightorange}\textbf{1.71$^{\pm 0.02}$} \\
& \mb & 100\% & 60.7$^{\pm 2.5}$ & 235.6$^{\pm 16.5}$ & 340.8$^{\pm 9.8}$ & 150.4$^{\pm 2.0}$ & 3.91$^{\pm 0.03}$ & 3.35$^{\pm 0.04}$ & 9.20$^{\pm 0.09}$ & 130.5$^{\pm 9.5}$ & 109.5$^{\pm 18.6}$ & 3.29$^{\pm 0.09}$ & 1.74$^{\pm 0.01}$ \\
& \mc & 100\% & 59.4$^{\pm 1.2}$ & 229.4$^{\pm 12.1}$ & 322.9$^{\pm 7.0}$ & \cellcolor{lightorange}\textbf{129.7$^{\pm 0.5}$} & 3.93$^{\pm 0.04}$ & 3.33$^{\pm 0.02}$ & 9.30$^{\pm 0.13}$ & 134.4$^{\pm 3.0}$ & \cellcolor{lightorange}\textbf{75.7$^{\pm 5.1}$} & 3.37$^{\pm 0.14}$ & 1.74$^{\pm 0.01}$ \\
& \md & 100\% & 70.3$^{\pm 1.9}$ & 198.9$^{\pm 28.8}$ & 404.3$^{\pm 3.4}$ & 175.9$^{\pm 3.1}$ & 4.39$^{\pm 0.04}$ & 3.43$^{\pm 0.03}$ & 9.96$^{\pm 0.10}$ & 198.9$^{\pm 28.8}$ & 145.1$^{\pm 24.1}$ & 3.31$^{\pm 0.16}$ & 1.84$^{\pm 0.02}$ \\
& \textbf{\ours} & 100\% & \cellcolor{lighterorange}58.5$^{\pm 1.2}$ & \cellcolor{lightorange}\textbf{163.1$^{\pm 31.0}$} & \cellcolor{lightorange}\textbf{314.4$^{\pm 8.0}$} & \cellcolor{lighterorange}131.8$^{\pm 4.0}$ & \cellcolor{lighterorange}3.90$^{\pm 0.08}$ & \cellcolor{lighterorange}3.32$^{\pm 0.02}$ & \cellcolor{lighterorange}8.94$^{\pm 0.26}$ & \cellcolor{lightorange}\textbf{99.0$^{\pm 11.3}$} & \cellcolor{lighterorange}80.4$^{\pm 9.4}$ & \cellcolor{lightorange}\textbf{2.93$^{\pm 0.08}$} & \cellcolor{lighterorange}1.73$^{\pm 0.03}$ \\
\midrule

\multirow{5}{*}{\makecell[l]{w/ Slope\\ + Stair}} 
& \ma & 60\% & 91.0$^{\pm 36.9}$ & 702.9$^{\pm 203.9}$ & 462.4$^{\pm 192.2}$ & 185.3$^{\pm 16.5}$ & 5.22$^{\pm 0.53}$ & 3.86$^{\pm 0.21}$ & 13.01$^{\pm 1.17}$ & 358.9$^{\pm 72.4}$ & 337.1$^{\pm 117.8}$ & 4.33$^{\pm 0.45}$ & 2.30$^{\pm 0.05}$ \\
& \mb & 60\% & 93.2$^{\pm 52.6}$ & 448.1$^{\pm 122.3}$ & 503.2$^{\pm 312.7}$ & 165.5$^{\pm 26.4}$ & 4.84$^{\pm 0.32}$ & 3.66$^{\pm 0.14}$ & 11.87$^{\pm 0.84}$ & 222.1$^{\pm 37.6}$ & 183.7$^{\pm 88.7}$ & 4.49$^{\pm 0.17}$ & 2.30$^{\pm 0.04}$ \\
& \mc & 100\% & \cellcolor{lighterorange}65.3$^{\pm 6.0}$ & 336.2$^{\pm 69.8}$ & 367.8$^{\pm 44.9}$ & \cellcolor{lighterorange}151.5$^{\pm 1.1}$ & \cellcolor{lightorange}\textbf{4.29$^{\pm 0.12}$} & \cellcolor{lightorange}\textbf{3.47$^{\pm 0.03}$} & \cellcolor{lightorange}\textbf{10.75$^{\pm 0.28}$} & 202.8$^{\pm 30.6}$ & \cellcolor{lighterorange}130.1$^{\pm 27.2}$ & \cellcolor{lighterorange}3.92$^{\pm 0.26}$ & \cellcolor{lighterorange}2.26$^{\pm 0.02}$ \\
& \md & 100\% & \cellcolor{lightorange}\textbf{62.3$^{\pm 6.9}$} & \cellcolor{lighterorange}320.3$^{\pm 129.0}$ & \cellcolor{lightorange}\textbf{321.0$^{\pm 61.0}$} & \cellcolor{lightorange}\textbf{144.1$^{\pm 8.2}$} & \cellcolor{lighterorange}4.63$^{\pm 0.44}$ & 3.83$^{\pm 0.05}$ & 11.92$^{\pm 1.25}$ & \cellcolor{lightorange}\textbf{149.7$^{\pm 5.6}$} & 179.1$^{\pm 173.9}$ & \cellcolor{lightorange}\textbf{3.53$^{\pm 0.63}$} & \cellcolor{lightorange}\textbf{1.72$^{\pm 0.47}$} \\
& \textbf{\ours} & 100\% & 67.7$^{\pm 1.9}$ & \cellcolor{lightorange}\textbf{250.1$^{\pm 29.4}$} & \cellcolor{lighterorange}327.8$^{\pm 7.5}$ & 154.2$^{\pm 4.2}$ & 4.75$^{\pm 0.16}$ & \cellcolor{lighterorange}3.56$^{\pm 0.07}$ & \cellcolor{lighterorange}11.98$^{\pm 0.61}$ & \cellcolor{lighterorange}157.2$^{\pm 3.8}$ & \cellcolor{lightorange}\textbf{121.4$^{\pm 7.8}$} & 4.01$^{\pm 0.02}$ & 2.29$^{\pm 0.01}$ \\
\bottomrule
\end{tabular}}
\end{table*}